\RequirePackage{fix-cm}

\documentclass{svjour3}                     
\smartqed  

\usepackage{natbib}
\setcitestyle{authoryear,open={(},close={)}}
\usepackage{subcaption}
\usepackage[utf8]{inputenc} 
\usepackage[T1]{fontenc}    
\usepackage{hyperref}       
\usepackage{url}            
\usepackage{booktabs}       
\usepackage{amsfonts}       
\usepackage{amsmath}
\usepackage{amssymb}
\usepackage{nicefrac}       
\usepackage{microtype}      
\usepackage{siunitx}
\usepackage{multirow}
\usepackage{breqn}
\usepackage[misc]{ifsym}
\usepackage{dsfont}
\usepackage{color}
\usepackage{tcolorbox}
\newcommand{\corrAuthor}{$^{\textrm{\Letter}}$}
\newcommand{\change}[1]{\textcolor{black}{#1}}

\newtheorem{deff}{Definition}
\usepackage{booktabs}
\usepackage{multirow}
\usepackage{tabularx}
\usepackage{algorithmic}
\usepackage[ruled,vlined]{algorithm2e}
\usepackage{listings}
\usepackage{enumitem}
\usepackage{bm}

\newlength\mylen
\newcommand\myinput[1]{%
  \settowidth\mylen{\KwIn{}}%
  \setlength\hangindent{\mylen}%
  \hspace*{\mylen}#1\\}
  
\newenvironment{algocolor}{%
   \setlength{\parindent}{0pt}
   \itshape
   \color{black}
}{}

 \usepackage{longtable}
 \usepackage{dashrule}

\journalname{Behavior Research Methods}

\begin{document}
\title{Automatic Discovery and Description of Human Planning Strategies}
\titlerunning{Automatic Discovery and Description of Human Planning Strategies}        

\author{Julian Skirzy\'nski\textsuperscript{1,2, }\corrAuthor \and Yash Raj Jain\textsuperscript{1} \and Falk Lieder\textsuperscript{1}
}

\authorrunning{J. Skirzy\'nski, Y. R. Jain, F. Lieder} 

\institute{\textsuperscript{1} Max Planck Institute for Intelligent Systems, T\"ubingen, Germany\\
           \textsuperscript{2} University of California, San Diego, CA 92093, USA\\
            \corrAuthor\email{julian.skirzynski@tuebingen.mpg.de} \\
}

\date{Received: date / Accepted: date}

\sisetup{tight-spacing=true}
\maketitle
\begin{abstract}
Scientific discovery concerns finding patterns in data and creating insightful hypotheses that explain these patterns. Traditionally, each step of this process required human ingenuity. But the galloping development of computer chips and advances in artificial intelligence (AI) make it increasingly more feasible to automate some parts of scientific discovery.
Understanding human planning is one of the fields in which AI has not yet been utilized. State-of-the-art methods for discovering new planning strategies still rely on manual data analysis. Data about the process of human planning is often used to group similar behaviors together. Researchers then use this data to formulate verbal descriptions of the strategies which might underlie those groups of behaviors.
In this work, we leverage AI to automate these two steps of scientific discovery. We introduce a method for automatic discovery and description of human planning strategies from process-tracing data collected with the Mouselab-MDP paradigm. Our method utilizes a new algorithm, called Human-Interpret, that performs imitation learning to describe sequences of planning operations in terms of a procedural formula and then translates that formula to natural language.
We test our method on a benchmark data set that researchers have previously scrutinized manually. We find that the descriptions of human planning strategies that we obtain automatically are about as understandable as human-generated descriptions. They also cover a substantial proportion of relevant types of human planning strategies that had been discovered manually. Our method saves scientists' time and effort, as all the reasoning about human planning is done automatically. This might make it feasible to more rapidly scale up the search for yet undiscovered cognitive strategies that people use for planning and decision-making to many new decision environments, populations, tasks, and domains. 
Given these results, we believe that the presented work may accelerate scientific discovery in psychology, and due to its generality, extend to problems from other fields.


\end{abstract}

\section{Introduction}
Scientific discovery is a product of scientific inquiry that allows generating and corroborating new insightful hypotheses. In the early days, scientific discovery has been seen as a prescriptive method for arriving at new knowledge, by gathering information and refining it with new experiments (see Bacon's \emph{Novum Organum} \citep{bacon1878} or Newton's \emph{Philosophiae Naturalis Principia Mathematica} \citep{newton1687}). The more established philosophical tradition argued that there exists a clear demarcation between the so-called context of discovery and the context of justification \citep{whewell1840,reichenbach1938,popper1935}. The former would be a product of logically unfathomable mental process: the "eureka" moment, that, if anything, could be rather studied by psychology. The latter would concern the proper verification and justification of the discovered theory, and would indeed have a formal structure. However, such an account leaves scientists at the mercy of having a eureka moment. This division was thus challenged. \citeauthor{kuhn1962} (\citeyear{kuhn1962}), for instance, saw scientific discovery as a complex process of paradigm changes, where increasing amounts of findings that disagree with the current paradigm lead to its change. 
Importantly for this work, this division was also challenged by early work on AI for problem-solving \citep{simon1971,simon1973}. Therein, discovery was understood as searching the problem space from the initial state representing current knowledge to a desirable goal state. The states transition from one to another by applying simple operators with a predefined meaning. The process of finding the shortest sequence of applied operators would create rules that could later be looked at by a human to determine what search heuristic has been used. The found search heuristic would determine the method for scientific discovery. Regardless of whether you agree with the distinction between discovery versus justification, these works showed that some steps involved in scientific discovery can be automated. Further work in such automation moved beyond the philosophical debate, with the main motivation becoming to aid scientists in their research \citep{addis2016computational}. 

In this article, we introduce a computational method for assisting scientists in studying human planning. Understanding how people plan is a difficult and time-consuming endeavor. Previous efforts to figure out which strategies and processes people use to make decisions and plan usually entailed manual analysis of the data \citep{payne1993adaptive,willemsen2011,CallawayLiederKrueger2017,Callaway2021}. The very act of finding the strategies has been thus left to researchers' ingenuity to discover the right patterns in the data. Moreover, previous work has largely failed to characterize people’s strategies in detail (but see \citep{jain2021computational,agrawal2020scaling,peterson2021using}). \change{Our method substantially simplifies that process by using an algorithm called Human-Interpret. Human-Interpret automatically discovers and describes human planning strategies externalized in process-tracing experiments. It achieves that by resorting only to a dictionary of logical primitives and their natural language counterparts that capture basic actions in those experiments. Human-Interpret imitates the input information gathering operations in terms of procedural logic formulas, and translates these formulas into natural language with the input dictionary. Our method runs Human-Interpret a number of times and selects among the strategies it found by applying a majority heuristic. To evaluate our method and, particularly, the Human-Interpret algorithm, we applied it to a planning task where people spend resources on inspecting nodes in a graph in order to find the most rewarding path to traverse. Our method generated descriptions of 4 human planning strategies for that task that were on par with the descriptions created through laborious manual analysis by \citep{jain2021computational}. Moreover, despite representing only a fraction of all strategies, the found strategies covered half of the relevant cases. Given these results, we believe that the presented work has the potential to facilitate scientific discovery in psychology and perhaps scientific discovery in general. Scientists can now use the help of AI not only for testing their hypotheses about how people make decisions but also for generating them. To apply our approach to new problems, such as multi-alternative risky-choice \citep{lieder2017automatic,peterson2021using}, it suffices to run new planning externalization studies and create a new dictionary of logical primitives.}

The outline of the article is as follows: We begin by providing background information and summarizing related work pertaining to our method and the benchmark problem used in Section 2. In the next section, we describe the whole pipeline of our method for the automatic discovery and description of human planning strategies. Section 4 shows the results of our test on the benchmark problem where we compare our automated pipeline with a standard manual approach. Lastly, Section 5 discusses opportunities for applying our method and directions for future work.

\section{Background and related work}
\change{In this section we detail how AI was used in scientific discovery as well as in scientific discovery for decision making and planning specifically.} Additionally, we present the methodology used to measure planning adapted in our studies and a state-of-the-art approach for discovering human planning strategies that we compared to. Lastly, we also describe an important part of our framework for automatic scientific discovery, that is an algorithm that describes planning strategies in an interpretable way by only using demonstrations of these strategies.

\subsection{Research on automating scientific discovery}
\change{One line of research involves the field of computational scientific discovery \citep{dvzeroski2007computational,sozou2017computational} which models the discovery problem mathematically and advances the discovery of laws or relations using artificial intelligence (AI). BACON \citep{langley1987scientific}, for instance, was a system for inducing numeric laws from experimental data. Having dependent and independent variables, it created taxonomies of these variables by clustering equally-valued dependent variables and defining new variables as products or ratios of independent variables. Then, \citeauthor{langley1983three} \citeyear{langley1983three}) created GLAUBHER that was formulating qualitative laws over the categories in the taxonomy, and STAHL which produced structural theories based on the data leading to anomalous behavior of existing theories (c.f. \citeauthor{kuhn1962} (\citeyear{kuhn1962})). \citeauthor{addis2016computational} (\citeyear{addis2016computational}) suggested to represent theories as programs and use genetic search for best theories. Other systems such as PHINEAS, COAST or IDS are described at length in a review paper by \citeauthor{shrager1990} (\citeyear{shrager1990}). An overview work by \citeauthor{dvzeroski2007computational} (\citeyear{dvzeroski2007computational}) shows an extension of this field to mathematical modeling: the background knowledge is represented as generic processes, the data takes form of time-series, and discovery takes place by deriving sets of explanatory differential equations.}

\change{Another line of research which used AI to help scientists in their endeavors was automatic experimental design. This approach follows the mathematical foundations of design optimization, in which the expected information gain of an experiment defines its usefulness in testing a hypothesis \citep{myung2013tutorial}. \citeauthor{vincent2017darc} (\citeyear{vincent2017darc}) used this approach to automate the creation of intertemporal choice experiments and risky choice experiments. \citeauthor{ouyang2016practical} (\citeyear{ouyang2016practical,ouyang2018webppl}) went a step further and created a system to automatically find informative scientific experiments in general. Their method expected a formal experiment space, expressed in terms of a probabilistic program as input and returned a list of experiments ranked by their expected information gain. The experiment that was the highest on that list would provably provide the most information to differentiate between competing hypotheses. \citeauthor{foster2019variational} (\citeyear{foster2019variational}) further refined the whole idea by introducing efficient expected information gain estimators.}

\subsection{Using artificial intelligence to understand human planning}
\change{The case for using AI in the quest of understanding human planning has been clearly made for one-shot decision-making scenarios \citep{agrawal2020scaling, bhatia2021machine, peterson2021using}. \citeauthor{agrawal2020scaling} (\citeyear{agrawal2020scaling}) fit simple models to the data and optimized them with respect to the regret obtained by comparing simple models' predictions to overly complex models' predictions. After the simple models' predictions converged to those of the complex models, they were used as a proper formalization of one-shot human decision strategies. \citeauthor{peterson2021using} (\citeyear{peterson2021using}) employed artificial neural networks to search for theories of one-shot decision-making. At first, they created a taxonomy of theories that expressed relations between available decision items (such as gambles, and whether the gambles are dependent or independent), and which covered the entire space of possible decision-making theories. Subsequently, they used neural networks to express those theories by imposing different constraints on those networks. The authors gathered a very large data set of human decisions and determined which theory is the best fit based on the networks' performance. Very recent works started going beyond one-shot decision making, and showed utility of machine learning for studying multi-step decision making, i.e. planning. We are aware of 2 such endeavors. \citet{fang2022machine} suggest to use the data set of human decisions and train machine learning classifiers to learn the association between features extracted from the data (e.g. reaction time), and the decision strategy provided as the label. Benchmark tests with SVM and KNN models showed that this approach is able to correctly discover strategies such as take-the-best. In the paper we compare to, \citet{jain2021computational} defined a computational method that assigns planning strategies to human decision data through Bayesian inference. Here, we go further than all the mentioned articles. Although we also consider the issue of i) discovering detailed human multi-step decision strategies (planning strategies), ii) we aim to discover those strategies from data automatically
without creating the initial model, whether a complex black-box model \citep{agrawal2020scaling}, a taxonomy of decision-making theories \citep{peterson2021using}, or a set of possible planning strategies \citep{fang2022machine, jain2021computational}.}

\subsection{Methods for measuring how people plan} \label{sec:processTracing}
People's planning for years has been an elusive process that lacked principled analysis tools. To study planning, psychologists firstly focused on one-step decisions and most often relied on educated guesses, i.e. self-constructed mathematical models of human behavior that captured the relationship between inputs and outputs of decision-making \citep{abelson1985,westenberg1994multi,ford1989process}. These methods were not error-proof because they sometimes fit conflicting models equally well to the same data or were too limited to capture the whole decision-making process \citep{ford1989process,riedl2008identifying}. To mitigate these drawbacks, scientists have developed \emph{process-tracing} methods that captured the process used in decision-making by also analyzing the context in which each of the steps were taken before reaching a decision \citep{payne1978exploring, svenson1979process}. Among multiple process-tracing paradigms, such as verbal protocols (e.g. in \citet{newell1972human}) or conversational protocols (e.g. in \citet{huber1997active}), some process-tracing paradigms were computerized and followed human choices by facing people with artificial tasks which required them to take a series of actions before making the final decision. One such process-tracing paradigm called the Mouselab \citep{payne1988adaptive} paradigm which is used for one-step decision making tasks was later adapted for studying planning under the name of Mouselab-MDP \citep{CallawayLiederKrueger2017} paradigm. Since we used this methodology in our framework for automatic scientific discovery for planning, we now introduce the Mouselab and Mouselab-MDP paradigms in more detail. At the end, we also introduce a state-of-the-art approach for scientific discovery for planning that uses Mouselab-MDP called the \emph{Computational Microscope}. Since it relies on manual analysis of data, we treat it as a baseline throughout this paper.

\subsubsection{Mouselab}\label{sec:mouselab}
The Mouselab \citep{payne1988adaptive} paradigm is one of the first computerized process-tracing paradigms. Its goal is to externalize some aspects of cognitive processes taking part in decision-making by engaging people in information acquisition that helps them reach a decision. Mouselab was developed to study multi-alternative risky choice. To do so, it presents participants with an initially occluded payoff matrix whose entries can be (temporarily) revealed by clicking on them. An $i,j$ entry of the $n\times m$ payoff matrix either hides a value for bet $i$ under outcome $j$, or, in the case of the first row, hides a probability for outcome $j$. The values are expressed in terms of dollars and they indicate the payoff a participant is expected to obtain after selecting gamble $i$ and having outcome $j$ to occur. Participant's task is to choose one of the available $i-1$ gambles based on the information they gathered by revealing the entries of the matrix. The sequence of clicks externalizes a participant's decision-making process by showing which information they considered. Using the Mouselab paradigm, scientists were able to discover that people choose strategies adaptively \citep{payne1988adaptive}. Despite its usefulness, however, Mouselab is inappropriate to study planning as selecting one gamble does not affect future gambles. Mouselab-MDP was developed to mitigate this shortcoming.

\subsubsection{Mouselab-MDP}\label{sec:mouselab-mdp}
\begin{figure}
    \centering
    \includegraphics[width=0.5\textwidth]{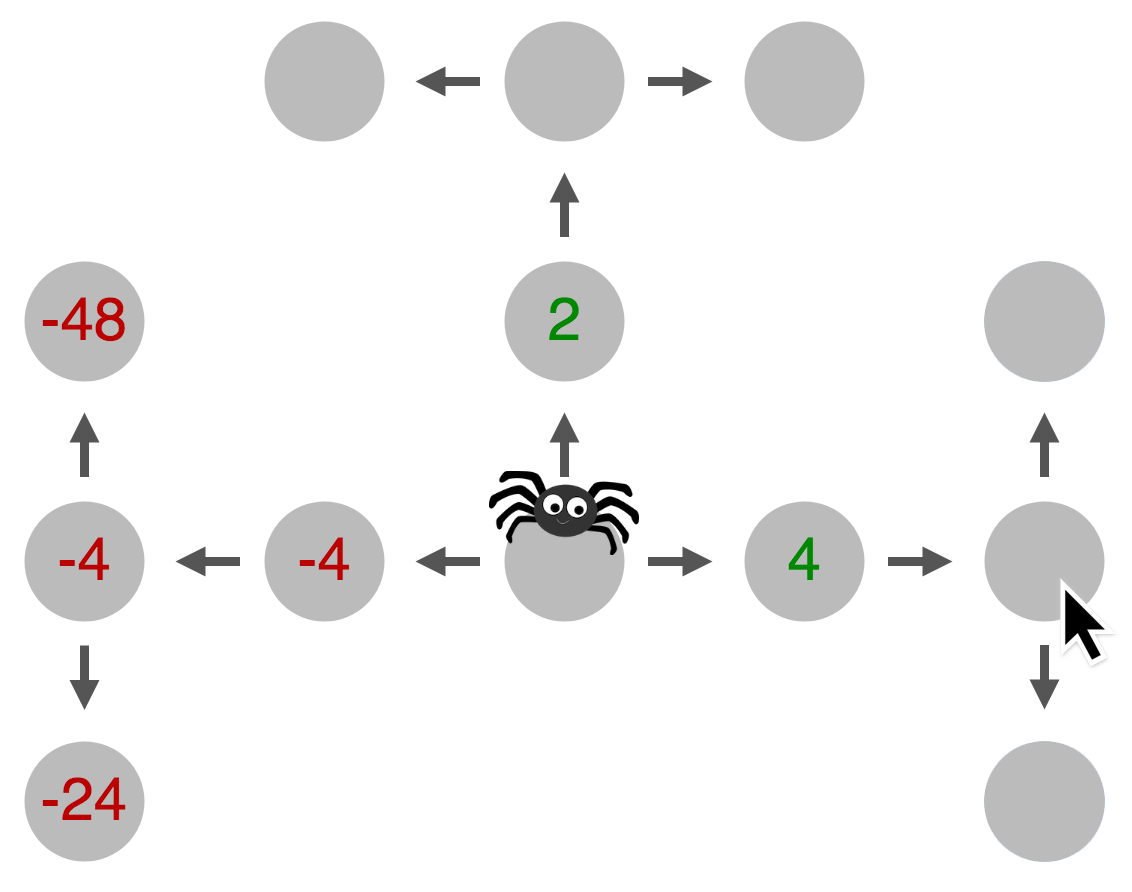}
    \caption{The Mouselab-MDP environment we used in the paper. The environment is a connected graph of nodes (grey circles) that hide rewards or punishments. The number hidden underneath a node can be uncovered by clicking on it and paying a small fee. The goal is to traverse a path starting in the black node and ending in a node at the highest level so that the sum of rewards along the way minus the cost of clicking was as high as possible.}
    \label{fig:mouselab-mdp}
\end{figure}

The Mouselab-MDP paradigm is a generalization of the Mouselab process-tracing paradigm in which participant's information-acquisition actions affect the availability of his or her future choices \citep{CallawayLiederKrueger2017}. By doing so and offering a way to externalize information-acquisition, Mouselab-MDP is suitable to study human planning. Concretely, a single decision about choosing a gamble is replaced with a Markov Decision Process (MDP; see Definition~\ref{deff:MDP}), and the payoff matrix is replaced with a graphical representation of the MDP, a directed graph with initially occluded nodes. 
\begin{deff}[Markov Decision Process]\label{deff:MDP}
A Markov decision process (MDP) is a tuple $(\mathcal{S},\mathcal{A},\mathcal{T}, \mathcal{R}, \gamma$) where $\mathcal{S}$ is a set of states; $\mathcal{A}$ is a set of actions; $\mathcal{T}(s,a,s') = \mathbb{P}(s_{t+1}=s'\mid s_t = s, a_t = a)$ for $s\neq s'\in\mathcal{S}, a\in\mathcal{A}$ is a state transition function; $\gamma\in (0,1)$ is a discount factor; $\mathcal{R}:\mathcal{S}\to\mathbb{R}$ is a reward function.
\end{deff}
Clicking on the node reveals a numerical reward or punishment hidden underneath it (see Figure \ref{fig:mouselab-mdp}). In the most commonly used setting of the Moueslab-MDP paradigm, a participant's goal is to find the most rewarding path for an agent to traverse from the start node to one of the terminal nodes (nodes without any out-going connections), by minimizing the number of clicks (each click has an associated cost). This formulation was used in a number of papers that studied human planning strategies \citep{griffiths2019doing,Callaway2021} and led to the creation of cognitive tutors that help people plan better \citep{CognitiveTutorsRLDM,CognitiveTutorsPNAS}, the creation of new, scalable and robust algorithms for hierarchical reinforcement learning \citep{kemtur2020leveraging,consul2021improving}, and helped in creating one of the first tools for analyzing human planning in more detail \citep{jain2021computational}.

\subsubsection{Categorizing planning strategies via the Computational Microscope}
The Mouselab-MDP paradigm gives information about the planning processes used by people \citep{Callaway2021,CallawayLiederKrueger2017}. Past research, similarly to the research on describing human decision-making, employed formal planning models (i.e. strategies) to describe the planning processes of people, and evaluated those models on data from experiments with human participants, e.g. \citep{botvinick2009hierarchically, Huys2012, huys2015interplay, callaway2018resource,Callaway2021}.
Researchers have come up with a number of planning models people could hypothetically use, including classic models of planning such as Breadth First Search, Depth First Search, models that use satisficing, etc. Recently, \citet{jain2021computational} created a new computational pipeline in which a set of manually created planning strategies are given as an input to a computational method, called the Computational Microscope, to fit them to human planning data. In more detail, the method takes process-tracing data generated using the Mouselab-MDP paradigm and by using Bayesian inference, categorizes the planning operations from trials of the Mouselab-MDP paradigm into a sequence of these planning strategies. One of its features is that, to categorize a trial, it incorporates all the information from all the trials before and after it. The authors of the paper performed manual inspection of all the process-tracing data (clicks) that participants made in each trial. They first found out similar clicks sequences based the trials and the features of the trial and then created generalized strategies that replicated these click sequences. This led to the creation of a set of 79 strategies.

One of the drawbacks of their method is the amount of time it took them to manually go through all the participants data and the scalability of their approach. Since our method aims to develop a method for automatic interpretation of the process-tracing data, which includes automatic discovery of the planning strategies, we treat their set of planning strategies and their approach as a baseline to compare our framework to.

\subsection{Finding interpretable descriptions of formal planning strategies (policies): AI-Interpret}
Part of our framework utilizes a variant of an algorithm developed to interpret reinforcement learning (RL; see Definition~\ref{deff:rl}) policies: AI-Interpret \citep{skirzynski2021automatic}. 
\begin{deff}[Reinforcement learning]\label{deff:rl}
Reinforcement learning (RL) is a class of methods that perform iterations over trials and evaluation on a given MDP in order to find the optimal policy $\pi^*$ which maximizes the expected reward  \citep{sutton2018reinforcement}. A deterministic policy $\pi$ is a function $\pi: \mathcal{S}\rightarrow\mathcal{A}$ that controls agent's behavior in an MDP and a nondeterministic policy $\pi$ is a function $\pi: \mathcal{S}\rightarrow Prob(\mathcal{A})$ that defines a probability distribution over the actions in the MDP. The reward $r_t$ represents the quality of performing action $a_t$ in state $s_t$. The cumulative return of a policy is a sum of its discounted rewards obtained in each step of interacting with the MDP, i.e. $G_t^{\pi}=\sum\limits_{i=t}^{\infty}\gamma^t r_t$ for $\gamma\in[0,1]$. The expected reward $J(\pi)$ of policy $\pi$ is equal to $J(\pi)=\mathbb{E}(G_0^{\pi})$.
\end{deff}
In contrast to existing methods for intepretability in RL that generate complex outputs: decision trees with algebraic constraints \citep{liu2018toward}, finite-state automata \citep{araki2019learning}, or programs \citep{verma2018programmatically} to represent policies, AI-Interpret generates simple and shallow disjunctive normal form formulas (DNFs; equivalent to decision trees, see Definition~\ref{def:dnf}) that express the strategy in terms of pre-defined logical predicates. 
\begin{deff}[Disjunctive Normal Form]\label{def:dnf}
Let $f_{i,j}, h: \mathcal{X}\rightarrow\{0,1\}, i,j\in\mathbb{N}$ be binary-valued functions (predicates) on domain $\mathcal{X}$. We say that $h$ is in disjunctive normal form (DNF) if the following property is satisfied:
\begin{align}
h(\boldsymbol{x}) = (f_{1,1}(\boldsymbol{x})\land...\land f_{1,n_1}((\boldsymbol{x}))\lor...\lor(f_{m,1}((\boldsymbol{x})\land...\land f_{m,n_m}((\boldsymbol{x})) \label{eq:3}
\end{align}
and $\forall i,j_1\neq j_2,\ f_{i,j_1} \neq f_{i,j_2}$. In other words, $h$ is a disjunction of conjunctions.
\end{deff}
Studies presented by \citet{skirzynski2021automatic} show that transforming this output into flowcharts, which use natural language instead of the predicates, is easily understood by people, and can even help in improving their planning skills. Moreover, AI-Interpret is an imitation learning method (see Definition~\ref{deff:il}) and interprets policies via their demonstrations. Due to these reasons, we decided to use it in order to achieve our goal: find descriptions of human planning strategies by using data from process-tracing experiments.
\begin{deff}[Imitation learning]\label{deff:il}
Imitation learning (IL) is the problem of finding a policy $\hat{\pi}$ that mimicks transitions provided in a data set of trajectories $\mathcal{D}=\{(s_i,a_i)\}_{i=1}^{M}$ where $s_{i}\in\mathcal{S}, a_i\in\mathcal{A}$ \citep{osa2018}.
\end{deff}

\change{On a high-level, AI-Interpret uses 4 inputs. If $\mathcal{S}$ is a set of states and $\mathcal{A}$ is a set of actions in a given environment, AI-Interpret accepts a data set of demonstrations (state-actions pairs) $\mathcal{D}=\{(s_i,a_i)\}_{i=1}^{N}, s_i\in\mathcal{S}, a_i\in\mathcal{A}$ generated by some policy $\pi$. Additionally, it also accepts the set of predicates $\mathcal{L}$ that act as feature detectors. Those feature detectors evaluate to $True$ or $False$ depending on the action to be taken and the current state, i.e.  $f: \mathcal{S}\times\mathcal{A}\rightarrow\{0,1\}$. On top of that, AI-Interpret uses a parameter $d$ denoting the maximum depth of the DNF (decision tree), and the ratio of the expected rewards $\alpha$. AI-Interpret uses $\mathcal{D}$ and $\mathcal{L}$ to find DNF formula $\psi$ of size at most $d$. Formula $\psi$ is required to induce policy $\pi_{\psi}$ with an expected reward of at least $\alpha$ of $\pi$'s expected reward. AI-Interpret achieves that by transforming each state-action pair in $\mathcal{D}$ into a vector of predicate valuations and clustering the set of these vectors into coherent groups of behaviors. In each iteration, the clustered vectors are used as positive examples for a DNF learning method and suboptimal state-action pairs generated by looking at possible actions in existing states, serve as negative examples. The DNF learning method is called Logical Program Policies (LPP; \citep{silver2020few}). LPP defines a prior distribution for the predicates in $\mathcal{L}$, and uses the MAP estimation and decision-tree learning methods to find the most probable DNF formulas that accept the positive examples and reject the negative examples. AI-Interpret uses LPP to find DNF $\psi$ that achieves expected reward defined by $\alpha$ and has depth limited by $d$. In case of a failure, it removes the least promising cluster (the smallest weighted posterior) to try describing the remaining data.}

\section{A new method for discovering and describing human planning strategies}
\begin{figure}
    \centering
    \includegraphics[width=\textwidth]{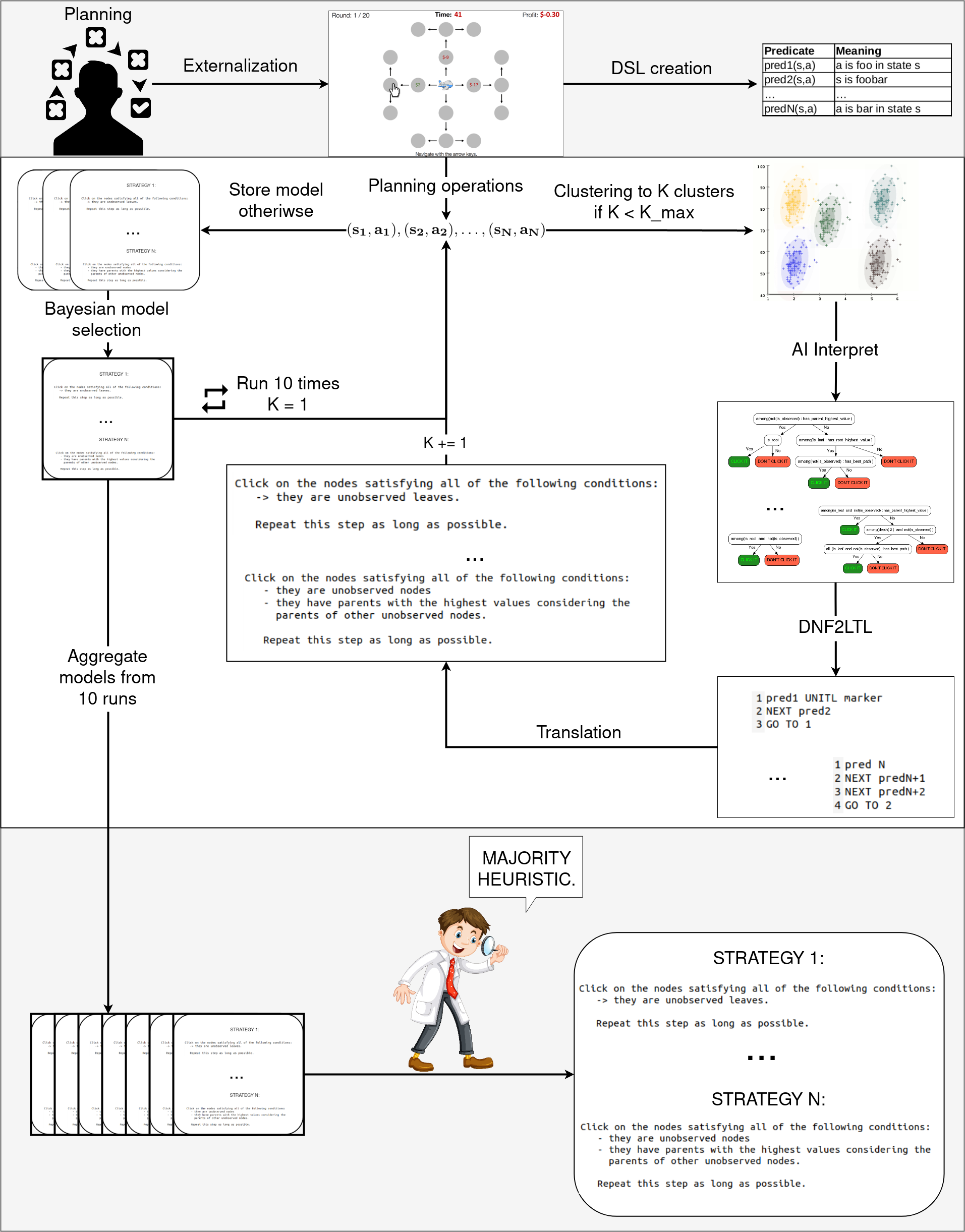}
    \caption{\change{Diagram representing our method for discovering and describing human planning strategies. The method assumes externalizing human planning first and gathering human planning operations in an experiment. Then it assumes creating a Domain Specific Language (DSL) to describe the environment in which participants made decisions. Afterwards, the method assumes running Human-Interpret, an algorithm that automatically extracts and describes possible strategies used by people. Human-Interpret starts by performing clustering on the gathered data to create mathematical representations of planning strategies -- policies. Then, it uses the AI-Interpret \citep{skirzynski2021automatic} algorithm supplied with the constructed DSL and the found policies to discover formulaic descriptions of the strategies. Later, it transforms the formulaic descriptions into procedural instructions expressed in linear temporal logic, and translates linear temporal logic formulas into natural language. Human-Interpret gradually increases the number of clusters and uses Bayesian model selection to output the best set of strategies it found. To generate the final result, our method assumes running Human-Interpret 10 times and applying a choice heuristic on 10 outputted sets.}}
    \label{fig:pipeline}
\end{figure}

We created a method that enables (cognitive) scientists to generate descriptions of human planning strategies using data from sequential decision-making experiments conducted with the Mouselab-MDP paradigm \citep{Callaway2021,CallawayLiederKrueger2017}. As illustrated in Figure~\ref{fig:pipeline}, our method comprises the following steps: 1) collecting and pre-processing process-tracing data on human planning, 2) setting up a vocabulary of logical predicates that can be used to describe people's strategies, \change{3) running our new algorithm 10 times to automatically discover strategies possibly used by people, 4) applying a choice heuristic to select which of those strategies are accurate. Importantly, during strategy discovery, our algorithm} automatically describes the found strategies as step-by-step procedures. The first four subsections describe each of these four steps in turn. We also detail the algorithm itself. The last section reports on the technical details of setting up the initial code base for our pipeline and can be skipped.

\subsection{Data collection and data preparation}\label{sub:data}
To use our method for discovering human strategies, the data collected in the experiments is required to meet certain criteria. Firstly, the experiment has to externalize people's planning by using a computerized process-tracing paradigms for planning (see ``Externalization'' in Figure~\ref{fig:pipeline}). In our benchmark studies we used the Mouselab-MDP paradigm \citep{Callaway2021,CallawayLiederKrueger2017}, since to our knowledge, it is the only such paradigm that is available so far. To study one-shot decision-making, on the other hand, one could use the standard Mouselab paradigm \citep{payne1988adaptive}, ISLab \citep{cook1993computerized}, MouseTrace \citep{jasper2002mousetrace} or other similar environments. Secondly, the experiments need to gather participants' \emph{planning operations} that we define as sequences of state-action pairs generated by each of the participants. The states are defined in terms of the information that the participant has collected about the task environment. The actions are the information gathering operations that the participant performs to arrive at their plan. Thirdly, the gathered planning operations should be saved in a CSV file that would additionally contain labels of the experimental block they came from (e.g. 'train' or 'test'), and labels corresponding to participant id. In our case this data was separated into distinct CSVs and custom Mouselab-MDP functions extracted states, actions, blocks and ids from them into a new (Python) object. The result of this process is visually presented in Figure~\ref{fig:pipeline} as the ``Planning operations'' arrow.

\subsection{Creating the Domain Specific Language}\label{sub:dsl}
Humans build sentences by using words (and build words by using phonemes). In order to automatically build descriptions of strategies used by people in the planning process-tracing experiments, we also need a vocabulary of some primitives. Due to the algorithmic setup of the method introduced in this section, here, the required set of primitives comprises logical predicates. Following \citeauthor{skirzynski2021automatic} (\citeyear{skirzynski2021automatic}) whose method is an important part of our pipeline, the predicates serve as feature-detectors and are formally defined as mappings from the set of state-action pairs to booleans, that is $f:\mathcal{S}\times\mathcal{A}\rightarrow\{0,1\}$. Practically, the predicates describe the state $s\in\mathcal{S}$, the action $a\in\mathcal{A}$ or the particular characteristic that action $a$ has in state $s$. For instance, predicate \texttt{is_observed(s,a)} might denote that node number $a$ in the Mouselab MDP paradigm has not yet been clicked and its value is hidden in state $s$. Later, this predicate could be used to define a very simple DNF that allows clicking all the nodes that have not yet been clicked, i.e. \texttt{not(is_observed(s,a))}. The second step in our pipeline thus considers creating a set of predicates that describe the process-tracing environment. Further in the text we will call that set of predicates the \textit{Domain Specific Language} (DSL). \change{Setting up the DSL is visually presented in Figure~\ref{fig:pipeline} as ``DSL creation'', and as we discussed, occurs after externalizing human planning.}

\change{As one can notice, creating an appropriate Domain Specific Language is a non-trivial task that is important to the success of the whole method. The process of creating the DSL usually entails studying the structure of the task and the strategies people are thought to use in that task. For instance, if we were to create a DSL for multi-alternative risky choice environment (such as Mouselab from Section~\ref{sec:mouselab}) we would first adhere to the existing knowledge of the various decision strategies people are thought to use in this task \citep{payne1993adaptive} and the elementary operations they are composed of \citep{bettman1990componential}. Inspecting those strategies would help us determine primitives needed to describe them, and these primitives would serve as initial predicates of the constructed DSL. In the case of risky choice, we would focus on characterizing information gathering operations. Furthermore, we would study the environment itself to expand the DSL with new primitives that described the properties of the environment. Those predicates could describe the gambles (e.g., most promising, second most promising, least explored, most explored, etc.) and the attributes (e.g., most informative one, one we know least about, etc.). By following this process, we would obtain a DSL with multiple predicates, some of which would stand for primitives known to be useful in describing people's planning heuristics, and some would denote characteristics of the environment that were yet unused. The algorithm's task would be to combine these primitives in one description so that they fit the sequences of planning operations externalized in the experiments.}

\subsection{Obtaining interpretable descriptions of human planning strategies via Human-Interpret}
The third step of our method assumes using the algorithm for discovering interpretable descriptions of people's planning strategies. This algorithm finds a set of planning strategies people used based on the data from process-tracing experiments, and then generates interpretable descriptions of those strategies in the form of procedural formulas. Since the AI-Interpret method of \citeauthor{skirzynski2021automatic} (\citeyear{skirzynski2021automatic}) plays a vital role in our pipeline, we call this new algorithm Human-Interpret.

\subsubsection{Big picture}
\change{Human-Interpret is an algorithm that discovers human planning strategies externalized in a process tracing paradigm in the form of natural language descriptions. As illustrated in Figure~\ref{fig:pipeline}, Human-Interpret clusters the participants' click sequences. It then applies the AI-Interpret algorithm to each cluster separately. AI-Interpret creates logical formulas that describe the planning strategies, and Human-Interpret translates those logical formulas into procedural descriptions in natural language. To maximize the likelihood of the final set of strategies obtained through this process, Human-Interpret also iterates through higher and higher number of strategies to discover, and eventually performs Bayesian model selection \citep{raftery1995bayesian} over the runs with respect to human data (see the transitions labeled ``Run 10 times" and ``Bayesian model selection" in Figure~\ref{fig:pipeline}). Technically, in each run, Human-Interpret executes 4 subroutines.}

\change{Firstly, Human-Interpret uses human data to generate demonstrations for AI-Interpret. It does so by clustering human data into input number of subsets represented as policies, and sampling those policies. Demonstrations obtained that way take form of representative sequences of information-gathering actions (clicks) paired with knowledge states in which they were performed (states) (see the transition labeled as ``Clustering to K clusters" that goes from the sequence of planning operations $(s_i,a_i)_{i=1}^N$ to the set of clusters in Figure~\ref{fig:pipeline}).}

\change{Secondly, Human-Interpret utilizes AI-Interpret to build descriptions of the sampled demonstrations. See the arrow labeled as ``AI-Interpret" in Figure~\ref{fig:pipeline} to visualize this process: AI-Interpret takes demonstrations from the clusters (middle left) and describes them using the created DSL (upper right) to output a number of flowcharts/DNF formulas describing planning strategies found in the demonstrations (middle right). The way in which these algorithms operate makes them robust to imperfections of the DSL. That is, even a highly incomplete DSL created according to the recipe from Section~\ref{sub:dsl} would likely result in approximate descriptions of some strategies. The reason is that AI-Interpret selects the subset of planning operations that are the easiest to describe with the existing DSL. Hence, although constructing the DSL does require human intelligence and domain knowledge, the robustness of our algorithm, makes creating an appropriate DSL simpler than it would otherwise be.}

\change{Thirdly, Human-Interpret enhances the interpretability of the found formulas using the method from \citep{becker2021} and turns the formulas into procedural descriptions mimicking the linear temporal logic formalism (see arrow ``DNF2LTL" in Figure~\ref{fig:pipeline} and consult Definition~\ref{def:ltl}).}
\begin{deff}[Linear Temporal Logic]\label{def:ltl}
Let $\mathcal{P}$ be the set of propositional variables $p$ (variables that can be either true or false), let $\lnot, \land, \lor$ be standard logical operators for negation, AND, and OR, respectively, and let $\mathbf{X}, \mathbf{U}, \mathbf{W}$ be modal operators for NEXT, UNTIL, and UNLESS, respectively. Linear temporal logic (LTL) is a logic defined on (potentially infinite) sequences of truth-assignments of propositional variables. LTL formulas are expressions that state which of the variables are true, and when they are true in the sequences. Whenever this agrees with the actual truth-assignment in an input sequence, then we say that a formula is true. 

Formally, for $\alpha$ and $\beta$ being LTL formulas, we define a formula to be expressed in LTL inductively: $\psi$ is an LTL formula if $\psi\in\mathcal{P}$ ($\psi$ states that one of the variables is true in the first truth-assignment in the sequence), $\psi= \lnot\alpha$ ($\psi$ is a negation of an LTL formula), $\psi=\alpha\lor\beta$ ($\psi$ is a disjunction of two LTL formulas), $\psi=\alpha\land\beta$ ($\psi$ is a conjunction of two LTL formulas), $\psi = \mathbf{X} \alpha$ ($\psi$ states that LTL formula $\alpha$ is true starting from the next truth-assignment in the sequence) or $\psi = \alpha \mathbf{U} \beta$ ($\psi$ states that LTL formula $\alpha$ is true until some truth-assignment in the sequence where LTL formula $\beta$ becomes true).
\end{deff}
\change{Fourthly, Human-Interpret expresses the found procedural descriptions in natural language. It does so by translating each predicate into a sequence of words, abiding by predicate to expression translations and syntax rules predefined in a special predicate dictionary. We represent this step by arrow ``Translation" in Figure~\ref{fig:pipeline}.}

The pseudo code for Human-Interpret can be found in the Algorithm~\ref{alg:human-interpret} box; the explanation of its parameters is shown in Table~\ref{tab:params}.

\begin{algorithm}
\begin{algocolor}
\SetAlgoLined
\SetKwInput{KwInput}{Input}
\SetKwInput{KwOutput}{Output}
\KwInput{Id of the experiment $exp\_id$;}
\myinput{Number of participants $num\_participants$;}
\myinput{Block $block$;}
\myinput{Max number of EM clusters $max\_num\_clusters$;}
\myinput{Bayesian criterion $criterion$;}
\myinput{Number of sequences generated by the EM clusters $num\_trajs$;}
\myinput{Domain specific language $\mathcal{L}$;}
\myinput{Features for the EM softmax models $features$;}
\myinput{Tolerance for the EM clustering $tolerance$;}
\myinput{Tolerance ratio for the EM clustering $change\_ratio$;}
\myinput{Maximum size of the DNF's conjunctions in AI-Interpret $interpret\_size$;}
\myinput{Tolerance for AI-Interpret $ai\_tolerance$;}
\myinput{Number of rollouts for AI-Interpret $num\_rollouts$;}
\myinput{Number of clusters for AI-Interpret $num\_ai\_clusters$;}
\myinput{Expert reward in the environment $expert\_rew$;}
\myinput{Maximum divergence from the expert $max\_div$;}
\myinput{Threshold for choosing the unless/until conditions $threshold$;}
\myinput{Allowed unless/until predicates $allowed\_predicates$;}
\myinput{Redundant predicates $redundant\_predicates$;}
\myinput{Predicate dictionary $\texttt{DICT}$}
\KwOutput{Set of procedural formulas $\mathcal{P}$;}
\begin{algorithmic}[1]
 \STATE $\mathcal{P}_{best} = []$
 \FOR{$K$ in 1:$max\_num\_clusters$}{
     \STATE $\mathcal{P} = []$
     \STATE $planning\_operations\leftarrow \text{load\_human\_data}(exp\_id,\ num\_participants,\ block)$
     \STATE $EM\_clusters\leftarrow \text{cluster\_human\_data}(planning\_operations,\ K$, \par\hspace{5.2cm}$ features,\ tolerance,\ change\_ratio)$
     \FOR{$c$ in $EM\_clusters$}{
       \STATE $demos\leftarrow \text{generate\_demonstrations}(c,\ num\_trajs)$
       \STATE $DNF\leftarrow \text{AI-Interpret}(demos,\ interpret\_size,\ ai\_tolerance,\ num\_rollouts,$\par\hspace{2.95cm}$num\_ai\_clusters,\ expert\_rew,\ max\_div,\ \mathcal{L})$
       \STATE $proc\_formula\leftarrow \text{DNF2LTL}(DNF,\ threshold,\ allowed\_predicates,$\par\hspace{3.9cm}$redundant\_predicates)$
       \STATE $pruned\_proc\_formula\leftarrow \text{prune}(proc\_formula)$
       \STATE $\mathcal{P}.\text{append}(pruned\_proc\_formula)$
       }
     \ENDFOR
     \STATE $\mathcal{P}\leftarrow \text{translate}(\mathcal{P}, \texttt{DICT})$
     \IF{$criterion(\mathcal{P}) > criterion(\mathcal{P}_{best})$}
     \STATE $\mathcal{P}_{best} = \mathcal{P}$
     \ENDIF
 }
 \ENDFOR
 \RETURN $\mathcal{P}_{best}$
 \end{algorithmic}
 \caption{Pseudocode for the Human-Interpret algorithm.}
 \label{alg:human-interpret}
\end{algocolor}
 \end{algorithm}

\begin{table}
\centering
\begin{tabularx}{\textwidth}{p{17mm}|p{27mm}|X} \toprule
\textbf{Algorithm} & \textbf{Parameter} & \textbf{Explanation} \\
\midrule
\multirow{6}{4em}{Human-Interpret} & exp\_id & Name of the experiment for which the clusters of state-action pairs will be created. The name is used to identify proper folders with the data. \\ 
& num\_participants & Number of participants whose data from the experiment to consider. \\ 
& block & Part of the experiment from which the data is taken (e.g. 'test'). Depends on which identifiers have been used in the experiment. \\ 
& max\_num\_clusters & Maximum number of probabilsitic (EM) clusters to create. \\
& criterion & Criterion for performing Bayesian model selection on models with differing number of clusters. \\
& num\_trajs & Number of sequences of planning operations generated by the EM clusters used to find the descriptions of the clusters. \\
& $\mathcal{L}$ & Domain Specific Language of predicates to create descriptions from. \\
& $\texttt{DICT}$ & Predicate to natural language expression dictionary. \\
\cmidrule{1-3}
\multirow{3}{4em}{Expectation-Maximization} & features & Set of functions which defines features in the environment for externalizing planning. The features are used by the softmax models that symbolize the EM clusters. Might be related or unrelated to the DSL. \\
& tolerance & Minimum change in the likelihood of the state-action sequences' assignment to EM clusters that allows further iterations of the EM algorithm. \\
& change\_tolerance & Minimum relative change between the likelihood in the previous and current iteration that allows further iterations of the EM algorithm.  \\
\cmidrule{1-3}
\multirow{3}{4em}{AI-Interpret} & interpret\_size, ai\_tolerance, num\_rollouts, num\_ai_clusters & See \citeauthor{skirzynski2021automatic} (\citeyear{skirzynski2021automatic}). \\
& expert\_reward & Average reward of the optimal policy in the environment \\
& max\_divergence & The maximum difference in formula-induced policy's expected reward and the mean reward of the interpreted policy, measured proportionally to the expert reward \\
\cmidrule{1-3}
\multirow{2}{4em}{DNF2LTL} & threshold & Proportion of demonstrations that might have a candidate until condition incompatible with the candidate conditions for other demonstrations, but still be selected as the until condition in the output formula. \\
& allowed\_predicates & Candidates for the until and unless conditions selected from the DSL. \\
& redundant\_predicates & List of unwanted predicates removed from the  DNF. \\
\bottomrule
\end{tabularx}
\caption{Explanation of the parameters used by Human-Interpret and other methods utilized by Human-Interpret.}
\label{tab:params}
\end{table}

\subsubsection{Examples}
\begin{figure}[ht]
    \centering
    \includegraphics[width=\linewidth]{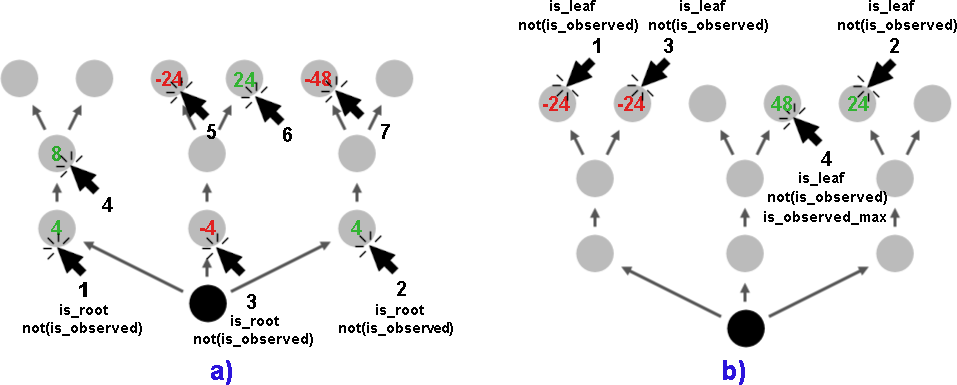}
    \caption{\change{Sample sequences of planning operations externalized in the Mouselab-MDP paradigm. Numbers below the click arrows denote the ordering of the planning operations (clicks), whereas the predicates written below the numbers denote relevant elements of the Domain Specific Language active when using a given planning operation.}}
    \label{fig:ex}
\end{figure}

\change{Consider the following two examples to better understand the workflow of Human-Interpret: Assume we are searching for planning strategies used by humans in the Mouselab MDP visible in Figure~\ref{fig:ex}a) and Figure~\ref{fig:ex}b).}
\change{The planning operations sequences we observed -- so the nodes clicked to uncover rewards -- are represented by numbers written next to the click arrows. The sample click sequences visible in the figures start with the node labeled as 1, the next element is the node labeled as 2, then 3, 4, etc. To relate to our pipeline shown in Figure~\ref{fig:pipeline}, the figures show planning operations externalized in a process tracing paradigm, the Mouselab-MDP. Assume we observed multiple click sequences as in Figure~\ref{fig:ex}a) and b), and the clustering step resulted in 2 policies that were able to reproduce them. The clicks sequences seen in the figures can thus additionally serve to visualize the output of the clusters from the clustering step in Figure~\ref{fig:pipeline}. Assume we further created a Domain Specific Language that includes predicates describing the nodes, and the attributes of the clicks on the nodes. Relevant predicates that evaluated to true when node number $i$ was clicked are written below $i$ in Figure~\ref{fig:ex}a) and Figure~\ref{fig:ex}b). As we can see, in Figure~\ref{fig:ex}a) the first 3 clicks activated predicate \texttt{is\_root}, i.e. they considered clicking the nodes of the MDP closest to the start, black node. Similarly, in Figure~\ref{fig:ex}b) the first clicks activated \texttt{is_leaf} predicate (nodes furthest to the black node), whereas predicate \texttt{is\_max\_observed} evaluated to $False$ for all but for the last node in the sequence, when a big reward was observed. In all cases, predicate \texttt{is\_observed} evaluated to $False$, i.e. each click was made on a node that has not been clicked yet. Human-Interpret uses multiple sequences of planning operations generated by the clusters alongside the predicates activated in each step of the sequences and runs AI-Interpret. AI-Interpret outputs a DNF formula for each cluster to capture the dynamics of the sequences. Assume that for Figure~\ref{fig:ex}a) the output was \texttt{(not(is\_observed) and is\_root and not(all\_roots\_observed)) OR (not(is\_observed) and all\_roots\_observed)}. $\newline$Since clicks other than the first 3 for sequences such as the one in Figure~\ref{fig:ex}a) activated many predicates, and only the predicate \texttt{not(is\_observed)} evaluated to $True$ for all of them, AI-Interpret found the best possible DNF formula that agreed with all the elements in the sequences. For cluster that generated sequences as the one in Figure~\ref{fig:ex}b), the found DNF formula was \texttt{not(is\_observed) and is\_leaf and not(is\_max\_observed)}. This DNF captures that the click sequences always started with clicking the leaf nodes and terminated whenever the maximum value was observed. In the next step, Human-Interpret uses an algorithm created by \citeauthor{becker2021} (\citeyear{becker2021}) to transform the DNF formulas into linear temporal logic formulas. For the provided DNF formulas, the output would be \texttt{not(is\_observed) and is\_root and UNTIL all\_roots\_observed THEN True UNTIL IT STOPS APPLYING} (the expression after \texttt{THEN} captures complete uncertainty) and \texttt{not(is\_observed) and is\_leaf UNTIL is\_max\_observed}. Finally, Human-Interprets translates the procedural descriptions into natural language by making use of a predefined dictionary. The final output we would get for our examples if we used the dictionary same as in this paper would be: \textit{1. Click on the nodes satisfying all of the following conditions: they are unobserved roots. Repeat this step until all the roots are observed. 2. Terminate or click on some random nodes and then terminate. Repeat this step as long as possible.} for the first cluster exemplified by the click sequence in Figure~\ref{fig:ex}a) and \textit{1. Click on the nodes satisfying all of the following conditions: they are unobserved leaves. Repeat this step until the previously observed node uncovers a 48.} for the second cluster exemplified by the sequence in Figure~\ref{fig:ex}b).}

\change{Human-Interpret would repeat these steps across all runs that consider different total number of clusters, say from 1 to 10, and employ Bayesian model selection to decide which set of strategies has the highest score under selected Bayesian criterion (e.g. marginal likelihood). This set would become the final output of the algorithm.}

\subsubsection{Clustering planning operations into planning strategies}
Human-Interpret begins the process with extracting planning strategies out of sequences of planning operations gathered in the process-tracing experiment. Let $\mathcal{D}=\{(\tau_{1i})_{i=1}^{K_{1}}, (\tau_{2i})_{i=1}^{K_{2}}\dots, (\tau_{Mi})_{i=1}^{K_{M}}\}$ denote the set of planning operations belonging to $M$ participants where $\tau_{ji} = ((s^{ji}_l,a^{ji}_l))_{l=1}^{L^{ji}}$ is the $i$-th sequence of planning operations generated by participant $j$, and $L^{ji}$ is its length. Human-Interpret utilizes the Expectation Maximization (EM) \citep{dempster1977maximum, moon1996expectation} algorithm to fit a probabilistic clustering model to the state-action sequences in $\mathcal{D}$ and extract $k$ planning strategies (the clusters $\pi_1, \pi_2 ...., \pi_k$). Each planning strategy corresponds to a softmax policy of the form given in Equation~\ref{eq:softmax}. 

\begin{equation}
    \pi_i(a\mid s, w_i) = \frac{\exp(\mathbf{f}(s,a)^{\intercal}w_i)}{\sum\limits_{k=1}^{T}\exp(\mathbf{f}(s,a_k)^{\intercal}w_i)} \label{eq:softmax}
\end{equation} 

Each softmax policy ($\pi_i$) is represented by weights \textbf{$w_i$} assigned to $P$ different features $\mathbf{f}=[\phi_1,\phi_2,\dots,\phi_P]$ of the state-action pair where $\phi_i: \mathcal{S}\times\mathcal{A}\rightarrow\mathcal{X}_{\phi}$. These features were partially derived from the DSL and partially hand-designed. There were 19 of them and they can be found in Appendix A.4. The aim of the EM algorithm is to find the planning strategies by clustering the click sequences in $\mathcal{D}$ into $k$ clusters, with each cluster being represented by some softmax policy $\pi_i$. It does this by optimizing the set of weights $W = (w_1, w_2, ...., w_k)$ that maximize the total likelihood ($\mathcal{M}$) of the click sequences under all the clusters. $\mathcal{M}$ is described in Equation~\ref{eq:EM}.

\begin{equation}
    \mathcal{M}(\mathcal{D} \mid W) = \sum\limits_{i=1}^{k} \sum\limits_{C \in \mathcal{D}} \sum\limits_{(s, a) \in C} \pi_i(a\mid s, w_i)
    \label{eq:EM}
\end{equation}

After obtaining the policies represented by the weights $W$, they are discretized to form new uniform policies ($\overline{\pi}_i$) as described in Equation~\ref{eq:discreteSoftmax} that assign uniform probability to actions with the highest probability according to $\pi_i$, that is to optimal actions. Policies $\overline{\pi_i}$ are then used to create a data set of demonstrations $\overline{\mathcal{D}}=\{(s_i,a_i)\}_{i=1}^L$ which contains $L$ planning operations generated through some fixed number of rollouts.

\begin{equation}
\overline{\pi}_i(a\mid s) = \begin{cases} 0, &\text{if }\pi_i(s,a) \neq \max\limits_{k\leq P} \pi(s,a_k)\\
                                      1/\lvert\{a_i: \pi_i(s,a_i) = \max\limits_{k\leq P} \pi(s,a_k)\}\rvert &\text{otherwise}
\end{cases}
\label{eq:discreteSoftmax}
\end{equation}

\subsubsection{Finding formulaic descriptions of planning strategies}
After computing policies $\overline{\pi}_i$ and generating the data set of demonstrations $\overline{\mathcal{D}}$, Human-Interpret essentially runs AI-Interpret \citep{skirzynski2021automatic}. The only modification to the AI-Interpret algorithm that is introduced by Human-Interpret relates to the fact that the input demonstrations no longer represent the optimal policy, but \emph{some}, often imperfect policy mimicking humans: $\overline{\pi}_i$ (further called the interpreted policy). Because of that, the expected reward for policies induced by candidate formulas cannot make at least $\alpha$ of the expected reward for the interpreted policy (see the Background section), since it leads to errors. For instance, in a situation when the interpreted policy achieves a reward of $R$, and the optimal policy achieves a reward of $100R$, a formula that induces a policy with reward $50R$ meets the criterion defined by $\alpha$, but is a poor approximation to $\overline{\pi}_i$. Because of that, Human-Interpret uses \emph{divergence} instead of the expected reward ratio. If $\pi_f$ is a policy induced by formula $f$ found by AI-Interpret and $\pi_{opt}$ is the optimal policy for the studied environment (here, the Mouselab MDP), Human-Interpret computes the divergence of $f$ as the ratio between the difference in rewards for the interpreted policy $\overline{\pi}_i$ and $\pi_f$, and the expected reward of the optimal policy, i.e. $\text{div}(\pi_f) = \frac{J(\pi_f) - J(\overline{\pi}_i)}{J(\pi_{opt})}$. Consequently, Human-Interpret searches for solutions whose size is limited by $d$ and for which the divergence is at most $\alpha$ (see the Background section). Note that introducing that modification requires the modeler, or the algorithm, to compute the optimal policy for the environment or at least know its approximation.

\subsubsection{Extracting procedural descriptions from logical formulas}
The output produced by the modified AI-Interpret algorithm we defined above is a DNF formula $f^*$. Following the finding that procedural descriptions are easier to grasp for people than flowcharts \citep{becker2021}, Human-Interpret uses the method presented in \citeauthor{becker2021} (\citeyear{becker2021}) to transform $f^*$ into a logical expression written in linear temporal logic. The DNF2LTL algorithm described in \citep{becker2021} produces such an expression by separating the DNF into conjunctions, and finding the dynamics of their changes in truth valuations using the initial set of demonstrations inputted to AI-Interpret. The output, which we will call a \emph{procedural formula}, separates the conjunctions with NEXT commands, and instructs to follow each conjunction until some condition occurs unless another condition occurs. The conditions are chosen among the predicates or 2-element disjunctions of predicates from the DSL introduced in Section~\ref{sub:dsl}, or simply read (for the UNTIL condition) ``until it applies'', which denotes a special logical operator. Since the output formula might be overly complex after that process, in the next step the algorithm prunes some of the predicates appearing in the conjunctions. Concretely, predicates are greedily removed one by one so as to increase the probability of people's planning operations under the strategy described by the shortened formula.

\subsubsection{Translating to natural language}
Once the procedural formulas are generated, it is possible to obtain fully understandable descriptions of human planning strategies by transforming the predicates and the operators appearing in the formulas into natural language. In our case, the procedural formulas are expressed in natural language by using a predefined predicate-to-expression dictionary that provides a direct translation for each predicate (as a sequence of words), and controls which words should be added, and in what order, to mimic the meaning of LTL operators.

\subsubsection{\change{Bayesian model selection}}\label{sec:bayes}
\change{Human-Interpret uses Bayesian model selection according to Bayesian inference with a uniform prior on the number of clusters \citep{kass1995bayes}. In more detail, it considers $K$ runs of human data clustering and cluster description, where output of run $i$ posits the existence of $i$ clusters and establishes model $i$. Each cluster for a model is defined as a mixture of two policies. The first policy is induced by the procedural description constructed for the cluster. The second policy serves as an error model. These two policies assign uniform probability to actions allowed and disallowed by the procedural description constructed for the cluster, respectively. The weights assigned to both policies are cluster-dependent free parameters with prior sampled using Beta functions where alpha and beta are hyperparameters (i.e., $\epsilon_i\sim\text{Beta}(\alpha, \beta)$ and $\epsilon_i\in[0,1]$). Mathematically, the clustering model $P(\tau)$ for a sequence of planning operations $\tau=(s_i,a_i)_{i=1}^T$, for $K$ clusters represented by $K$ policies $\pi_1,\dots,\pi_K$, and for $K$ error models $\overline{\pi_1},\dots,\overline{\pi_K}$ took the following form: \begin{align}P(\tau\mid \bm{\epsilon}) & = \sum\limits_{i=1}^K 1/K * P_i(\tau; \epsilon_i) \\ & = \sum\limits_{i=1}^K 1/K * \prod\limits_{j=1}^T\left(\epsilon_i * \overline{\pi_i}(s_j,a_j) + (1-\epsilon_i) * \pi_i(s_j,a_j)\right).\end{align} 
Human-Interpret selects the best model using the equations above according to input Bayesian criterion: either the marginal likelihood, the Bayesian Information Criterion (BIC) \citep{konishi2008information}, or the Akaike Information Criterion \citep{vrieze2012model}.}

\subsection{\change{Heuristically choosing strategies used by people}}
\change{In the last step of our method, we aggregate the results of 10 runs performed by Human-Interpret, i.e. 10 Bayesian-optimal sets of strategies, and select the final output. In order to do that, we employ the majority heuristic. According to the heuristic, a strategy belongs to the output, if it was discovered in at least 7 of the models/runs. Otherwise, it is most likely noise. Our heuristic uses 7 as the majority criterion since it was found to maximize the accuracy of our method during evaluation (see Section~\ref{sec:eval}).}

\subsection{Technical details regarding using our method}
Here, we present technical details connected to installing and using our method for discovering and describing human planning strategies. We equipped the initial code base written in Python 3.6 with 1) data from 4 planning experiments ran in different versions of the Mouselab-MDP paradigm, 2) a Domain Specific Language (DSL) of logical primitives used to generate procedural formulas, and 3) a dictionary of predicate-to-expression entries for transforming a formula into natural language. Each of these elements is either a parameter or a hyperparameter of Human-Interpret. A description of the experiments can be found in \citep{jain2021computational}, whereas the DSL is detailed in Appendix A.2 and contained in one of the files in the code base, similarly to the dictionary. Thanks to the initial values for those parameters, it is possible to use our method without performing prior research. Moreover, one may extend our research by slightly modifying the DSL or running Human-Interpret on a different data set. The steps involved in setting up the whole method are as follows:
\begin{enumerate}
    \item Download data needed in the pipeline and the source code for Human-Interpret by cloning the appropriate Github repository using the command:
    \begin{center}
\texttt{git clone https://github.com/RationalityEnhancement/\\InterpretableHumanPlanning.git}
    \end{center}
The repository includes four data sets that are contained in the folder \texttt{data/human}. Refer to \citet{jain2021computational} for a detailed description of the experiments they come from.
    \item Access the root directory of the downloaded source code and install the needed Python dependencies:
        \begin{center}
\texttt{pip3 install -r requirements.txt}
    \end{center}
    \item Run Human-Interpret on either of the available data sets by typing
    \begin{center}
        \texttt{python3 pipeline.py ---run <r> ---experiment\_id <name> ---max\_num\_strategies <max> ---num\_participants <num\_p> ---expert\_reward <exp\_rew> ---num_demos <demos> --begin <b>}
    \end{center}
    Parameter \texttt{run} is the run id of the call to the function. Then, \texttt{experiment\_id} corresponds to the name of the experiment that resulted in one of the four data sets. As written in Table~\ref{tab:params}, \texttt{max\_num\_strategies} quantifies the maximum number of strategies that could exist in the data set and should be described; parameter \texttt{expert\_reward} defines the maximum reward obtainable in the Mouselab MDP defining each of the experiments; \texttt{num\_participants} state that the data of the first \texttt{<num>} participants should be extracted from the data set; \texttt{num\_demos} controls how many strategy demonstrations to use in Human-Interpret; \texttt{begin} controls which model to begin with (how many clusters to consider in the first model to then incrementally increase that number thus defining consecutive models). The available names, number of all tested participants, and corresponding expert rewards are provided in the readme file included in the source code. The output of this command is saved in the \texttt{interprets\_procedure} folder in a text file named according to the following structure (and smaller files without the prefix): $$\texttt{BEST\_strategies\_<name>\_<max>\_<num\_p>\_<demos>\_run<run>}.$$ The files contain procedural formulas describing the clusters, their natural language descriptions, and a set of statistics associated with those descriptions (e.g. how well they fit the experimental data, how big they are, etc.)
    \item To reproduce the exact set of strategies from this paper, run the following code for \emph{i} in 1:10:
    \begin{center}
        \texttt{python3 pipeline.py ---run i ---experiment v1.0 ---max\_num\_strategies 17 --begin 17 ---num\_participants 0 ---expert\_reward 39.97 ---num_demos 128}
    \end{center}
\end{enumerate}

Information regarding other parameters available for tuning the Human-Interpret algorithm can be found in Table~\ref{tab:params} and in file \texttt{pipeline.py}. The set of logical primitives serving for our DSL may be accessed by navigating to \texttt{RL2DT/PLP/DSL.py}. Finally, the dictionary is included in file \texttt{translation.py}. To apply our method to different problems consult \texttt{pipeline.py}, \texttt{README.md}, Section~\ref{sub:data}, and Section~\ref{sub:dsl}.

\section{Evaluating our method for discovering human planning strategies}
\change{In this section, we evaluate the reliability our method on data from a planning externalization experiment that utilized the Mouselab-MDP paradigm. We measure reliability threefolds. First, we analyze the quality of all the strategies that Human-Interpret discovered automatically. Second, we measure the observational error of the final set of strategies outputted by our method with respect to strategies discovered through laborious manual inspection of the same data set. Third, we compare the performance of our computational method do the manual method. The first subsection focuses on the setup of our method: it describes the planning experiment, the vocabulary of logical primitives and the parameters for Human-Interpret. The second subsection numerically describes all of the discovered strategies. The third subsection measures the observational error of our method. The last subsection compares our method to the manual method.}

\subsection{Setup of the benchmark problem}
\subsubsection{Planning experiment}\label{sec:exp}
In our benchmark problem, we used a process-tracing experiment on human planning conducted according to the Mouselab-MDP paradigm (see Section~\ref{sec:mouselab-mdp}), namely the first experiment presented in \citeauthor{jain2021computational} (\citeyear{jain2021computational}). This experiment (which we will refer to as the increasing variance experiment) used the Mouselab-MDP environment shown in Figure~\ref{fig:mouselab-mdp} where the rewards hidden underneath the nodes were different in every trial but were always drawn from the same probability distribution. The nodes closest to the spider's starting position at the center of the web had rewards sampled uniformly from the set $\{-4,-2,2,4\}$. Nodes one step further away from the spider's starting position had rewards sampled uniformly from the set $\{-8,-4,4,8\}$. Finally, the nodes furthest away from the starting position of the spider harbored rewards sampled uniformly from the set $\{-48,-24,24,48\}$. The fee for clicking a node to uncover its reward was $1$. The 180 participants who took part in this experiment were divided into 3 groups that differed in what kind of feedback was provided during the trials. The control group received no feedback. The second group received feedback on their first move. The third group received feedback on every click they made that was designed to teach them the planning strategy that is optimal for the task environment. There were always 2 blocks: a training block and a test block, with 20 trials in the training block and 10 trials in the test block. As stated in the previous sections, the experiment utilized the Mouselab-MDP process-tracing paradigm and operationalized people's planning strategies in terms of the clicks (planning operations) they performed.

\subsubsection{Domain Specific Language (DSL) and translation dictionary}\label{sec:dsl}
We adopted the DSL of predicates from our work on AI-Interpret which was also conducted on Mouselab-MDP environments \citep{skirzynski2021automatic}. Generally, the DSL consisted of over 14000 predicates generated according to a hand-made context-free grammar. The predicates were defined on state-action pairs, where states were represented as Python objects capturing the uncovered and covered rewards in the Mouselab-MDP, and each action denoted the ID of the node to be clicked. The predicates described the current state of the Mouselab-MDP or the actions available in that state. The state is described in terms of the clicked nodes, the termination reward, and other properties. The actions are described in graph theoretic terms, such as the depth of the clicked node, whether it is a leaf node, whether its parents or children have been observed, and so on. The DSL included two crucial second-order predicates: the \emph{among} predicate asserts that a certain condition (given by a first-order predicate) holds among a set of nodes defined by another first-order predicate, and the \emph{all} predicate asserts that all the nodes satisfying a certain condition also satisfy another condition. Detailed descriptions of the predicates in the DSL we used are available in Appendix A.2.

The dictionary we used for translating the resulting procedural formulas is an adapted version of the dictionary from \citeauthor{becker2021} (\citeyear{becker2021}). In our dictionary, we changed natural language translations of most predicates to use graph theoretic jargon (such as leaves, roots, etc.). Moreover, our translations always begin with ``\emph{Click on the nodes satisfying all the following conditions:}'' if there are any non-negated predicates in the formula and list the non-negated predicates as conditions. The difference with the original dictionary is also that more complex predicates (such as those which included other predicates as their argument) are broken down into 2 or more conditions, whereas in the original dictionary each predicate has its unique translation. More details on how the translation was created can be found in our project's repository in the \texttt{translation.py} file.

\subsubsection{Parameters for the Human-Interpret algorithm}
To run Human-Interpret, we used the default parameters for the AI-Interpret algorithm, the DSL described in the previous section, the data from all of the participants in the experiment described in Section~\ref{sec:exp} (that is 180), and data from both blocks of the experiment (i.e., the training block and the test block). We ran Human-Interpret with 1-20 clusters and had it perform model-order selection according to the BIC. The model selection was set to disregard clustering models for which Human-Interpret was unable to produce a description for all clusters. Our rationale for doing so was that models that do not describe all the clusters are not useful for the purpose of understanding human planning. The DSL we used and the predicates we used for the DNF2LTL algorithm can also be found in Appendix A.2 and A.3. All parameters are listed in Table~\ref{tab:params2} in Appendix A.1. The more elusive parameters, such as interpret\_size or num\_trajs were selected based on the simulations and interpretability experiments presented by \citet{skirzynski2021automatic}.

\subsection{Measuring the reliability of Human-Interpret}

\subsubsection{Method}
\change{To assess the reliability of Human-Interpret, we ran clustering and description subroutines of Human-Interpret 21 times to output 21 maximum BIC models, and analyzed each subset of 10 models with respect to the ground truth, i.e. the strategies found by \citet{jain2021computational}. In total, the runs found 21 distinct strategies (found in Appendix A.5), where 10 of the 21 strategies were rediscovered with respect to the manual analysis by \citet{jain2021computational}. Since every strategy outputted by Human-Interpret is represented in terms of a cluster of planning operations and the policy-mixture describing the cluster, we measured the average of the the mean likelihood per planning operation in the clusters under the policy mixtures assigned to the unique strategy. For comparison purposes, we additionally measured the likelihoods under two additional methods: the random method represented by a policy that always assigns the same probability to all possible planning operations (including termination), and the Computational Microscope \citep{jain2021computational} whose policies are programmatic versions of the strategies \citet{jain2021computational} discovered manually.}

\subsubsection{Results}
\begin{figure}
    \centering
    \includegraphics[width=\textwidth]{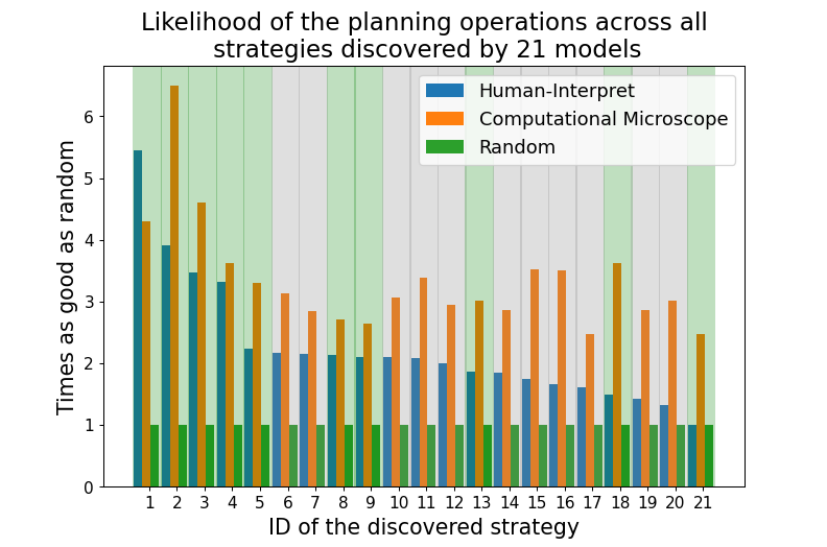}
    \caption{\change{Improvement over the random strategy in terms of the standard mean likelihood per planning operation for 3 methods: Human Interpret, the Computational Microscope and a random method that always assigns equal probability to all possible actions. The listed strategies are those found in the initial 21 strategy models. The bars show improvement in terms of the mean likelihood per planning operation. Background colors encode whether a strategy was chosen as ground truth (green) or not (grey).}}
    \label{fig:improvement}
\end{figure}

\change{Figure~\ref{fig:improvement} shows the aggregated results of the runs by reporting the average improvement in the mean likelihood over the random method of both Human-Interpret and the Computational Microscope across all the discovered strategies. As it can be seen, the planning operations are over twice as better fitted to the strategies that represent them than if they were assigned at random. Unsurprisingly, the same planning operations are also better fitted to the manually found strategies from the Computational Microscope than to the ones automatically discovered by Human-Interpret.}

\subsection{Measuring the reliability of our computational method}\label{sec:eval}
\change{
\subsubsection{Method}
The output of our method depends on how the outputs of the 10 runs of Human-Interpret are aggregated. The majority heuristic used to perform this aggregation has one parameter: the number of runs that have to agree on a discovered strategy. We therefore evaluated the reliability and reproducibility of our method using cross-validation. This involved splitting the set of all 10-run subsets into a training set that contained 70\% of all 300 thousand subsets and a test set that contained the remaining 30\% (about 100 thousad subsets). We selected the value of the parameter by maximizing the following accuracy metric on the training set:
\begin{align}\label{eq:acc}
    acc(S) &= \frac{\vert b\in B: b\in G\vert + \vert b\notin B: b\notin G\vert}{|S|},
\end{align}}
\change{where $G$ is the ground truth, $S$ is the set of all strategies, and $B$ is the set of strategies discovered by our computational method. The parameter value that optimized the method's accuracy on the training set was 7.}

\change{We then evaluated the method's performance with the chosen parameter value on the test set. We measured the method performance using four standard metrics from signal detection theory: accuracy, precision, recall, and F1 score \citep{mcnicol2005primer}. A method's precision is the proportion of strategies discovered by our method that correspond to the ground truth strategies, whereas recall defines the proportion of the total number of the ground truth strategies our method discovers. Formally, the precision $prec$, recall $rec$ and $F1$ measures are defined as
\begin{align}
    prec(S) &= \frac{\vert b\in B: b\in G\vert}{|B|}, \\
    rec(S) &= \frac{\vert b\in B: b\in G\vert}{|G|}, \\
    F1(S) &= 2 \cdot \frac{prec(S)\cdot rec(S)}{prec(S) + rec(S)}.
\end{align}
We chose F1-score as the secondary measure of reliability because it emphasized the number of true positives. Lastly, we also measure the stability of our method by estimating the proportion of runs that output the median result we report here, and its variability that quantifies how many strategies are in the difference set if 2 outputs are different.}

\subsubsection{Results}
\change{
As summarized in Table~\ref{tab:stability}, our method has high precision of 80\%, moderately high accuracy of 67\%, and rather low recall of 40\%. The F1-score of 0.53 indicates that the method presents good reliability. The stability of our method is equal to 79.5\%, meaning that there is roughly an 80\% chance of generating the same results when the method is run twice. If the results are different, however, they differ by 1.45 strategies on average. We will refer to this statistic as the method's \textit{variability}. The most common example of non-zero variability in our tests was that the smaller set was missing one strategy or that each set contained one strategy that was missing from the other one. The standard errors are mostly null, meaning that all these estimates are robust. Jointly, these numbers mean that around 80\% out of all strategies discovered by our computational method are actually used by people in planning, and these strategies represent slightly below a half of all the ground truth strategies Human-Interpret can currently discover. Using the example of the median run of our method with respect to the F1-score, it discovers 5 strategies where 4 are actually used by people, and 1 represents noise. The stability and the variability of the method indicate this is a representative output. We discuss how relevant the discovered strategies are in the next section.}

\begin{table}[ht]
\centering
{\color{black}\begin{tabular}{lllllll} \toprule
Run & F-1 score & recall & precision & accuracy & stability & variability \\
\midrule
Mean & $0.53 \pm 4e^{-5}$ & $0.8 \pm 2e^{-4}$ & $0.4 \pm 0$ & $0.67 \pm 0$ & $0.795 \pm 1e^{-3}$ & $1.45 \pm 1e^{-3}$\\
Median & 0.53 & 0.8 & 0.4 & 0.67 & -- & --\\
\bottomrule
    \end{tabular}}
    \caption{Error statistics for the output of our method with respect to the manually discovered ground truth. The mean run represents a run with average statistics, whereas the median run represents a run with median values for all the statistics. The numbers after the $\pm$ sign indicate standard errors.}
    \label{tab:stability}
\end{table}

\subsection{Evaluating and comparing strategies discovered by our computational method to strategies discovered through manual inspection}
\begin{figure}[ht]
    \centering
    \includegraphics[width=0.9\linewidth]{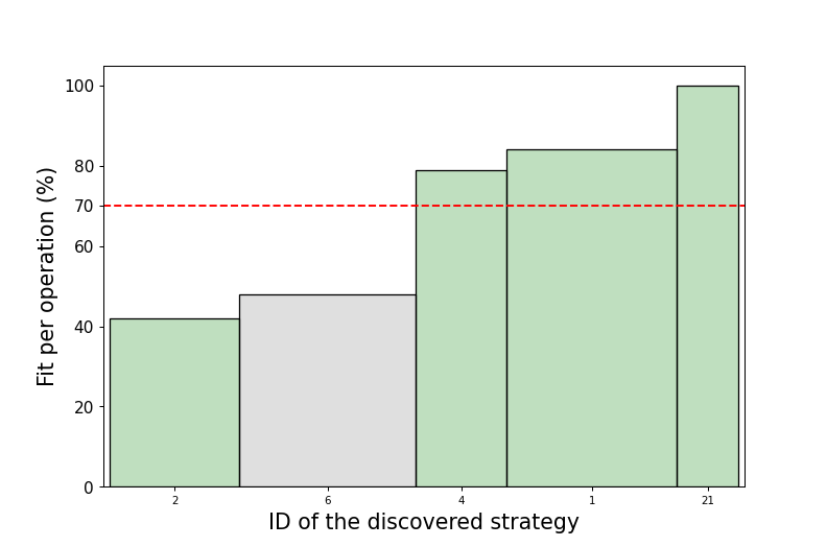}
    \caption{\change{Plot showing the quality of strategies found by our method, measured as the fit per operation (FPO), depending on their size. Strategies are labeled on the x-axis via their unique ID (see Table~\ref{tab:strategies} for a reference). The width of the bar corresponds to the size of the cluster representing the strategy. The height of the bar corresponds to the FPO measure. The red line indicates the average FPO across all clusters. Green color indicates strategies that belong to the ground truth, whereas grey color indicates noise.}}
    \label{fig:barplot}
\end{figure}

In this section we show the descriptions of human planning strategies discovered in the median run of our method, list their statistics, and compare them to the strategies that \citet{jain2021computational} found through manual inspection. In this and following sections we address the discovered strategies as Strategy $i$ for $i\in[21]$ relating to the numeration introduced in Figure~\ref{fig:barplot}. Some statistics on the strategies are reported with the standard error.

\change{\paragraph{Ground truth strategies} Human-Interpret outputted descriptions for 5 strategies seen in Table~\ref{tab:strategies}: Strategies 1, 2, 4, 6, and 21. Four of those strategies concur with the ground truth; only Strategy 6 is a false positive. The 4 rediscovered strategies jointly account for 51.3\% of people's planning operations \citep{jain2021computational}. Exactly 2 of the 4 strategies: Strategy 2 and Strategy 4, are mentioned as one of the most frequent according to \citet{jain2021computational}. There is one more frequent strategy that our method did not discover. However, our analysis of Strategy~1 indicates that  manual analysis heavily underrepresented its frequency \citet{jain2021computational}. It is because the average fit per planning operation (FPO) is significantly higher under this strategy than under the strategy assigned by the Computational Microscope. Since Strategy 1 is a special case of Strategy 4, and Strategy 4 has a frequency of over 36\% in the manual analysis, we suspect that many planning operations assigned to Strategy 1 were labeled as Strategy 4 by the manual analysis. Hence, Strategy 1 was underrepresented despite the fact it seems to be often used by people.}

\change{\paragraph{Noisy strategies} When it comes to the noisy strategies not represented in the ground truth, our method returned 1 such output. This output represents a particular behavior of Human-Interpret, namely an oversimplification of the strategies caused by the data selection process. Strategy 6, despite being very similar to one of the manually found strategies, lacks an early termination constraint. It was discovered in this form due to the operation selection process implemented by Human-Interpret (through AI-Interpret) where certain planning operations are removed from the set to simplify creating a description. Relevant data was hence likely removed and the strategy eventually became too specific: it considered only a subset of the planning operations. Since the strategy was chosen even after running Human-Interpret multiple times and applying the majority heuristic, the removed data must have been indescribable with the current DSL.}

\paragraph{Quantitative analysis of the strategies} Here, we report descriptive statistics about the complexity of the automatically generated descriptions, the frequencies of the discovered strategies, how well they corresponded to the EM clusters, and how well they explain the sequences of planning operations they are meant to describe. These statistics suggest that our method enables to find reasonable solutions of a very promising quality, especially considering that our method discovered those strategies without any labelled training examples or human feedback. On average, the discovered descriptions \change{agreed with the softmax models of the clusters on $84\% \pm 8$\%} of the planning operations. That quantity, which we call the \emph{formula-cluster fit} (FCF) is defined as the average of two proportions. The first one is the proportion of planning operations generated according to the inferred description that agreed with the choices of the corresponding softmax model. The second one is the equivalent proportion obtained by generating the demonstrations according to the softmax models and then evaluating them according to the descriptions. In both cases we performed 10000 rollouts. Further, the average likelihood of the planning operations within \change{the clusters reached $70\% \pm 11$\%} of the likelihood that would have been achieved if all planning operations perfectly followed the descriptions. We call that proportion the \emph{fit per operation} (FPO). The quality of individual cluster-descriptions, as measured in terms of the FPO, is depicted in Figure~\ref{fig:barplot}.
Since the average measurements over all clusters do not take into account the importance of clusters, we decided to perform the same computations using cluster size-dependent weighted averages. As larger clusters represent the most often used strategies, and these are the strategies we are predominantly interested in from a psychological standpoint, weighted averages capture the measured statistics with higher accuracy. After weighting our statistics \change{we found that the descriptions achieved a similar formula-cluster fit (FCF) of $86\% \pm 4\%$, and the fit per planning operation (FPO) of $66\% \pm 3\%$. We also tried excluding the unique cluster describing the random policy, which trivially achieves the FPO of 1, which caused the average to drop to $62\% \pm 3\%$. Those results suggest that the more accurate estimate of the fit with respect to planning operations is closer to $65\%$, whereas the fit with respect to the formulas is indeed around $85\%$}.

\newpage
{\color{black}\begin{longtable}[H]{p{0.03\textwidth}p{0.48\textwidth}p{0.07\textwidth}p{0.04\textwidth}p{0.04\textwidth}p{0.04\textwidth}p{0.03\textwidth}p{0.02\textwidth}} \toprule
\multirow{2}*{ID} & \multirow{2}*{Strategy descriptions}& \multicolumn{6}{c}{Statistics} \\
\cmidrule{3-8}
& & FR & FCF & FON & FPO & C & N \\
\midrule
\endhead
\multirow{11}*{4} & \textbf{Description: }
\begin{enumerate}
    \item Click on the nodes satisfying all of the following conditions: \begin{enumerate}[label=-]
        \item they are unobserved leaves. 
    \end{enumerate} 
    Repeat this step until all the leaves are observed or the previously observed node uncovers a 48.
\end{enumerate} & \multirow{2}*{16.7\%} & \multirow{2}*{0.88} & \multirow{2}*{0.86} & \multirow{2}*{0.75} & \multirow{2}*{5} & \multirow{2}*{10} \\
& \textbf{Summary: }Explore final outcomes until observing +48 & & & & & & \\
& \textbf{Manual counterpart: }Search for the best possible
final outcome & 36.6\% & & & & & \\
\cmidrule{1-8}

\multirow{23}*{1} & \textbf{Description: }
\begin{enumerate}
    \item Click on a node satisfying all of the following conditions: \begin{enumerate}[label=-]
        \item it is an unobserved leaf.
    \end{enumerate} 
    \item Unless the previously observed node uncovers a 48, in which case stop at the previous step, click on a node satisfying all of the following conditions: 
    \begin{enumerate}[label=-]
        \item it is an unobserved leaf.
    \end{enumerate}
    click in this way under the condition that:
    \begin{enumerate}[label=-]
        \item the previously observed node was its sibling.
    \end{enumerate}
    \item GOTO step 1 unless all the leaves are observed or the previously observed node uncovers a 48.
\end{enumerate} & \multirow{2}*{31\%} & \multirow{2}*{0.98} & \multirow{2}*{0.89} & \multirow{2}*{0.77} & \multirow{2}*{13} & \multirow{2}*{10} \\
& \textbf{Summary: }Explore final outcomes until observing +48 sibling by sibling. & & & & & & \\
& \textbf{Manual counterpart: }Explore final outcomes with preference for nodes in the same sub-tree of the root & 0.36\% & & & & & \\
\cmidrule{1-8}

\multirow{8}*{21} & \textbf{Description: }
\begin{enumerate}
    \item Terminate or click on some random nodes and then terminate. Repeat this step as long as possible.
\end{enumerate} & \multirow{2}*{4.3\%} & \multirow{2}*{0.53} & \multirow{2}*{1.0} & \multirow{2}*{1.0} & \multirow{2}*{1} & \multirow{2}*{10}\\
& \textbf{Summary: }\text{Random planning.} & & & & & & \\
& \textbf{Manual counterpart: }Random planning. & 1.2\% & & & & & \\
\cmidrule{1-8}

\multirow{3}*{2} & \textbf{Description: }Do not click. & \multirow{2}*{22.6\%} & \multirow{2}*{1.0} & \multirow{2}*{0.46} & \multirow{2}*{0.26} & \multirow{2}*{1} & \multirow{2}*{9}\\
& \textbf{Summary: }\text{No planning.} & & & & & & \\
& \textbf{Manual counterpart: }\text{No planning} & 13.2\% & & & & & \\
\cmidrule{1-8}

\multirow{16}*{6} & \textbf{Description: }
\begin{enumerate}
    \item Click on a node satisfying all of the following conditions: \begin{enumerate}[label=-]
        \item it is an unobserved leaf. 
    \end{enumerate} 
    \item Click on the nodes satisfying all of the following conditions: \begin{enumerate}[label=-]
        \item they are unobserved leaves. 
    \end{enumerate}
    Click in this way as long as: \begin{enumerate}[label=-]
        \item the previously observed node was their sibling. 
    \end{enumerate}
    Repeat this step as long as possible.
\end{enumerate} & \multirow{2}*{6.3\%} & \multirow{2}*{0.73} & \multirow{2}*{0.61} & \multirow{2}*{0.36} & \multirow{2}*{7} & \multirow{2}*{9} \\
& \textbf{Summary: }Explore all final outcomes sibling by sibling & & & & & & \\
& \textbf{Manual counterpart: } -- & -- & & & & & \\
\bottomrule
\caption{The strategies discovered by applying our computational method to the benchmark problem. The strategies are listed along side their ID, their automatically generated description, the summary we created by hand, and the name (or numbers) of the corresponding strategy (strategies) discovered manually by \citet{jain2021computational}. FR denotes the frequency of the strategy; FCF (fit cluster-formula) averages two proportions: formula demonstrations agreeing with the softmax clusters and vice-versa measured using 10000 demonstrations; FON (fit optimal-non-optimal) quantifies how often people's planning operations in the cluster agreed with the description; FPO (fit per operation) is the ratio between the average likelihood per planning operation belonging to the cluster and the average likelihood per planning operation for the (policy induced by the) cluster's description; C, which stands for complexity, is the number of individual predicates in the description; N is the number of clusters with the given description. Frequency (FR) is the only statistic measured for the manually found strategies. The statistics are averaged across all models in the median run of our method.}
\label{tab:strategies}
\end{longtable}}

\change{\paragraph{Quantitative comparison against the manual method} Generally, our method outputted 5 strategies that cover $51\%$ of the ground truth planning operations and describe them almost 3 times better than chance. In the case of the manual inspection performed by \citet{jain2021computational}, the results were higher, and the average improvement was almost 4 times as high as than the random method across 79 strategies. It has to be noted however, that our method analyzed only up to 20 clusters (thus strategies) contrary to \citet{jain2021computational}. Hence, the strategies discovered through our method were more coarse. Despite that limitation, our method managed to achieve a good accuracy score of 67\%, all within mere 20 days, which included problem analysis and DSL creation (around 2 weeks), running the code (around 5 days), and analyzing the results (around 1 day). In contrast, it took \citeauthor{jain2021computational} (\citeyear{jain2021computational}) about 120 days to finish their manual analysis. This means that our method could have accelerated the process of strategy discovery by over 3 months -- 100 days -- compared to manual inspection. Jointly, these results suggest that our method can be used to speed-up research on the strategies of human planning and decision-making. 
A more detailed summary of the comparison between the performance of our automated strategy discovery and description method with the manual approach by \citet{jain2021computational} can be found in Table~\ref{tab:comp}.}

\begin{table}[ht]
\centering
{\color{black}\begin{tabular}{lll} \toprule
& \multicolumn{2}{c}{Method} \\
\cmidrule{2-3}
Statistic & Automatic analysis & Manual analysis \\
\midrule
Runtime & \textbf{20 days} & 120 days \\
Discovered strategies & 5 & 79 \\
Accuracy & 0.67 & \textbf{1} \\
Success rate & 81\% & \textbf{100\%} \\
Ground truth represented & 51\% & \textbf{100\%} \\
Likelihood per click & $0.31 \pm 0.06$ & {$\mathbf{0.43 \pm 0.04}$} \\
Times as good as random & $2.99 \pm 0.7$ & {$\mathbf{3.93 \pm 0.61}$} \\
\bottomrule
    \end{tabular}
    \caption{Comparison between our method for automatically finding and describing human planning strategies and the manual approach by \citeauthor{jain2021computational} (\citeyear{jain2021computational}). Runtime is the number of days it took to generate the result. Discovered strategies counts how many strategies were generated the method. Accuracy measures the accuracy of the method according to Equation~\ref{eq:acc}. Success rate is the proportion of sequences of planning operations that the method eventually described. Ground truth represented measures the ground truth proportion of planning operations that are represented by the ground truth strategies discovered by the method. Likelihood per click is the average likelihood per a planning operation. Times as good as random quantifies the improvement in average likelihood per operation over the random model which assigns equal probability to all actions that are possible in a given step.}
    \label{tab:comp}}
\end{table}

\section{General discussion}
The main contribution of our research is automating the process of scientific discovery in the area of human planning. By using our method, scientists are no longer left at the mercy of their own ingenuity to notice and correctly interpret the right patterns in human behavior; instead they can rely on computational methods that do so reliably. Concretely, we developed the first method for the automatic discovery and description of human planning strategies. Our method's pipeline comprises 4 steps: The first step is to run an experiment that externalizes human planning operations. The second step is to create a Domain Specific Language (DSL) of logical predicates that describe the task environment and the planning operations. The third step is to run the generative imitation learning algorithm that we created and present in this paper, called Human-Interpret. Human-Interpret discovers the strategies externalized in the experiment by creating generative softmax models (the generative step) and describes them by procedural rules formulated in the DSL by imitating their rollouts (the imitation learning step).  \change{Finally, the fourth step is to apply a heuristic on the output produced by Human-Interpret to choose the final set of strategies.}

\subsection{Advantages}

Human-Interpret is the first-ever automatic method that can discover that people use previously unknown planning strategies and what they are. This is a step towards leveraging artificial intelligence to facilitate scientific discovery in research on human planning and decision-making. The main benefits of this approach are that it is more objective, potentially capable of discovering strategies that human researchers might overlook, and that it can be applied to many large data sets that human researchers do not have the capacity to analyze manually. The following paragraphs summarize the strengths of our method in more detail.

\paragraph{Reliability} The tests we ran on a benchmark planning problem revealed that the reliability of our method is comparable to that of a manual human analysis when it comes to the most frequent strategies.
\change{Firstly, the automatically discovered strategy descriptions were clear and understandable (see Table~\ref{tab:strategies}), mimicking the rigor and clarity of man-made descriptions.
Secondly, the overall accuracy of our method was good, reaching 67\% while keeping recall as high as $80\%$. This means that the strategies found by our method were highly likely to be actually used by people.
Finally, our automated approach found 2 out of 3 most important, frequently used strategies assuming the division from \citep{jain2021computational}, and provided evidence to label one additional strategy as relevant. The one remaining important strategy was not expressable with the current DSL (we explain that in paragraph \textit{Imperfect DSL}), and hence was not outputted. These results indicate that our method discovered almost the same set of important strategies as scientists did despite working with almost no human supervision.
Reliability of our method with respect to less frequent strategies likely requires improvements in the created DSL, as well as running Human-Interpret with many more clusters.}

\change{\paragraph{Time complexity} Applying our computational method was much faster than manual human analysis. It took us approximately 6 days before we obtained the final output via Human-Interpret and heuristic analysis of its output, whereas studying human planning without the aid of AI took about 120 days. We hence sped up the whole process 20 times. If we additionally included the time needed to set up a proper DSL (refer to Section~\ref{sec:dsl} to see how it was created),  the total time would accrue to 20 days. This is still a 6-fold improvement over the manual analysis. We also expect that for new problems, the reported time connected to creating the DSL would not be exceedingly different. The reason is that by building a DSL for the Mouselab-MDP we already cover a wider variety of problems: the problems we can represent concern deliberate decisions that involve selecting, processing, and integrating multiple pieces of information. Since that description applies to the majority of problems in the domain of planning, our DSL may serve as a basis for their study, and be simply extended. As extending existing work is much faster than creating something new, we think this process would not last longer than as in our case. When it comes to the time required to run Human-Interpret itself, it could differ depending on the size of the considered problem.\footnote{For a more detailed discusion of this issue, see the \textit{Computational complexity} paragraph of the next section.} In general, however, Human-Interpret can currently handle reasonably sized Mouselab-MDPs and if one is ready to sacrifice the possibility of discovering a larger number of strategies by selecting fewer clusters, it can even run in less than 3 days. Analyzing the output of Human-Interpret takes no longer than 1-2 days, irrespective of other parameters.}

\change{\paragraph{Generality} As we mentioned in the previous paragraph, our method could, in principle, be extended to a wide range of (sequential) decision problems. For instance, consider playing chess. If we wanted to inspect human strategies in chess, the process-tracing paradigm would have to record the sequence of moves and countermoves the player is considering while deciding what to do next. To achieve this, we could provide the player a second chessboard and ask them to use it to play through what might happen depending on which move they choose next. The belief state of the metalevel MDP would encode the sequence of moves and positions the player has considered up to a given time. The predicates would encode features that predict how uncertain the player should be about alternatives moves’ potential. This would include how often they have simulated playing the move, how many steps they looked ahead, and how much the most recent simulations have changed the player’s estimate of the move’s quality \citep[cf.][]{russek2022time}. The design of those predicates could also be informed by how grand masters represent chess positions \citep[e.g.,][]{chase1973perception,mcgrath2021acquisition}. The experiment would ask chess players to use the second chessboard to select which moves to analyze in what order before selecting the best move on the first chessboard. We could then apply Human-Interpret to their externalized planning operations. Human-Interpret would learn interpretable strategies that procedurally described which moves to think about, depending on how they are related to the current position and the player’s previous planning operations. We might thereby discover the clever planning strategies that enable chess players to efficiently identify excellent moves.}

\paragraph{Summary} \change{The main benefit of using the introduced pipeline is that it is more objective than visual inspection. Moreover, from a long-term perspective, our automatic method is a promising step towards leveraging artificial intelligence to accelerate scientific discovery in research on decision-making and planning. Rather than trying to manually discover one strategy at a time, our method makes it possible to run many large experiments with many different environments and discover the most representative strategies people use across those environments at once. A scientist interested in human planning could then invest his or her resources to other components of their research while waiting for the computations to finish. Afterwards, they could either start inspecting the data in search of more detailed strategies or use the found strategies as hypotheses to test in experimental studies. Further, our method is more objective than the subjective and potentially biased manual approach. Strategies and their descriptions are assigned based on mathematical likelihood and they are provably optimal under such probabilistic criteria, whereas people could introduce strategies based on their preconceptions and imperfect knowledge. Computers can tirelessly apply our rigorous, systematic procedure to all trials of all participants of numerous data sets. Our systematic, objective method might thereby be able to discover important strategies that scientists might otherwise overlook; one example thereof is Strategy~1 in Table~\ref{tab:strategies} which likely belongs to the set of frequently used strategies, but the manual analysis incorporated most of it to its generalization (Strategy 4).}

\subsection{Limitations}
\change{A major limitation of our method is that it was unable to rediscover all the strategies types documented by \citet{jain2021computational}. In this section, we discuss the underlying bottlenecks and other limitations of our method}

\change{\paragraph{Imperfect clustering} The softmax clusters did not always contain planning operations that fit well together. For instance, the No planning Strategy 2 had a poor fit score of 0.26 with respect to people’s planning operations (FPO), although the fit between the description and the cluster softmax model (FCF) was virtually perfect (1.0). This means the description was an accurate representation of the Human-Interpret cluster and the poor FPO came from incompatible planning operations. Most likely, the softmax strategies could not capture the underlying logic of the clusters, as planning operations clustered together by the generative part of Human-Interpret were too diverse.}

\change{\paragraph{Incomplete DSL} Our analysis indicates that the DSL we used was incapable of capturing all of the strategies, which was the most evident in the case of the 3rd most important, frequent planning strategy from \citet{jain2021computational} called ``Consecutive second maximum strategy''. This strategy was used in around 6\% of the trials of the planning experiment.} It observes the final outcomes in a random order but stops planning after it encounters two outcomes with the second-highest reward consecutively. The reason it was not discovered by our method was because the DSL did not include a predicate that described the second-highest reward.

\paragraph{Disregarding evidence for some planning strategies} Table~\ref{tab:strategies} shows differences in the coverage (FR) between the same strategies found by our pipeline and manual analysis. The main discrepancy is that the automatically discovered strategies have a higher frequency than the human-discovered strategies. Besides Strategy 1 and 4 which we believe to be improperly represented by the manual analysis, those differences stem from the simplification inherent to our method and used by AI-Interpret -- it finds a description that fits some subset of the elements. Note that since the softmax clusters represent human planning operations, when a demonstration (planning operation) of the softmax cluster is rejected by the algorithm, a human operation that corresponded to this demonstration (e.g. it followed the same planning principle) is also rejected. A rejection of a cluster-generated planning operation in AI-Interpret might occur in two situations: 1) either it is indescribable due to the imperfect DSL, or 2) it differs from other operations generated by the softmax cluster due to the imperfect clustering. In fact, we computed that AI-Interpret utilized only 67\% of the cluster-generated planning operations to find the descriptions. A part of human planning operations were thus also rejected when finding a description, but still counted towards the coverage of the strategy. Based on our measure of the quality of the discovered descriptions with respect to human planning operations, i.e. the FPO which was as high as \change{70\%, the upper bound for the number of human planning operations which were completely disregarded by Human-Interpret is probably around 30\%.} This issue might be solved in future work by studying more clusters and improving the DSL.

\change{\paragraph{Computational complexity}
One of our method's limitations is that it requires a considerable amount of compute time. Because the required amount of time increase with the maximal number of clusters considered by the method, the method's computational complexity is a bottleneck to how many strategies can be discovered within a given amount of time. From the preliminary runs we tried and other gathered data, we suspect it would take about 4 times longer to run our code with up to 80 clusters. As much as this is still manageable, our experiments served mainly as a proof of concept, and thus we decided for fewer clusters. In fact, we believe that even though we did not manage to find strategies that are as fine-grained as the manually found ones due to computation time concerns, Human-Interpret can be scaled to a larger number of clusters. In our initial runs, the code took about 40 days before terminating. After optimizing our code using hashing and parallelization, the required amount of compute time dropped to 5 days. The remaining computational bottlenecks concern i) clustering sequences of planning operations via the EM algorithm (up to 2.5 days), ii) evaluating all the predicates in the DSL on all the states from the demonstrated sequences (up to 1 hour per cluster), iii) computing DNF formulas based on this evaluation (up to 1 hour per cluster). Further code optimization in those areas may improve the time complexity of the method to an even larger degree. However, our method’s computational complexity will likely remain a limiting factor for some practical applications. }

\change{\paragraph{Bias towards fewer strategies.}
We strove for a balance between parsimony and expressiveness of our strategies by introducing the complexity parameters (preferring smaller decision trees in AI-Interpret, limiting their size). It is possible that by foregoing these parameters, we could discover a larger number of strategies. They could, however, be less interpretable, not always distinct, and sometimes try to explain the randomness inherent to some human decision-making.}

\subsection{Future work}
\paragraph{Tackling the limitations} As stated above, the main limitations of our method are: imperfect softmax clustering, an incomplete DSL, and the computational bottlenecks of AI-Interpret. We could improve the clustering by imposing a no-diversity penalty on the models. Hypothetically, this would make the softmax models as different as possible and they could thus capture more diverse kinds of behaviors, for instance as in \citep{eysenbach2018diversity}. Alternatively, we could obtain a better clustering by incorporating a different method for choosing which number of softmax clusters is optimal. Here, we used the maximum marginal likelihood or the highest BIC to differentiate between competing models (understood as numbers of clusters), but there might be better ways for accounting for model complexity, such as performing cross-validation. We see improving the DSL as an iterative refinement process in which we would add the necessary predicates to the DSL, check the results of running our computational method with the new DSL against the manual analysis, identify missing predicates, if any, and then return to the first step. Just a few iterations of this process should render a DSL capable of describing most of the strategies used by people in the Mouselab-MDP environment. Both of those enhancements, however, would not be feasible without first making our computational method run faster. Thus, upgrading the implementation of AI-Interpret should be one of the first next steps.

\paragraph{Merging Human-Interpret with the Computational Microscope} Besides ameliorating the performance of our method, a worthwhile future work direction is to combine the strategy discovery method presented in this article with our computational process-tracing method for measuring which planning strategy each participant used on each trial of the experiment -- the Computational Microscope \citep{jain2021computational}. We hypothesize that our method could serve to establish the basic set of strategies one could use as input to the microscope. Then, a human planning sequence could be automatically assigned to one of the automatically discovered strategies, instead to the manually discovered strategies. In this way, automatic strategy discovery might make it possible to speed up the development of equivalent computational process-tracing methods for other tasks and domains and to improve the repertoire of strategies that those methods use to describe the temporal evolution of people's cognitive strategies.

\paragraph{Researching other scenarios} Finally, future work should seek to apply our method for the automatic discovery and description of human strategies to other scenarios. The methodology we presented in this paper can be easily extended to decision-making in other domains, such as risky choice, intertemporal choice, and multi-attribute decision-making \citep{lieder2017automatic}). \change{Our method could thereby help distill the existing process-tracing data on these tasks (Mouselab and eye-tracking) into detailed process models of the specific heuristics people use to make those kinds of decisions. It would likely lead to the discovery of new heuristics and a more mechanistic understanding of some of the heuristic processes that are already known. It would also help decision researchers go beyond studying individual decision strategies to automatically identifying the entire toolbox of all strategies that decision makers have available. Our method could additionally be used to accelerate the slow process of strategy discovery because it could be simultaneously applied to process-tracing data from many different types of decision environments. In that way, we could also gain insights into how the types of heuristics people use differ across different environments.} Because our approach is rather general, we believe it has the potential to accelerate scientific discovery in several areas of the cognitive and behavioral sciences.

\section*{Declarations}

\paragraph{Funding}
This project was funded by German Federal Ministry of Education and Research (BMBF): Tübingen AI Center, FKZ: 01IS18039B.

\paragraph{Conflicts of interest/Competing interests}
The authors declare that they have no conflicts of interest or competing interests.

\paragraph{Availability of data and material (data transparency)}
Anonymized data from the experiment we used as our benchmark experiment is available at \url{https://github.com/RationalityEnhancement/InterpretableHumanPlanning/tree/main/data/human}.

\paragraph{Code availability (software application or custom code)}
The code for Human-Interpret is available at \\ \url{https://github.com/RationalityEnhancement/InterpretableHumanPlanning}.

\paragraph{Ethics approval}
Not applicable
\paragraph{Consent to participate (include appropriate statements)}
Not applicable
\paragraph{Consent for publication}
Not applicable.

\bibliography{main}  
\newpage
\appendix
\renewcommand{\thesection}{A.\arabic{section}}
\renewcommand{\thefigure}{A\arabic{figure}}
\setcounter{table}{0}
\renewcommand{\thetable}{A\arabic{table}}
\section*{\Large{Appendix}}  

\section{Human-Interpret parameters}
Table~\ref{tab:params2} lists the values of the parameters of the Human-Interpret method that we used in our benchmark planning environment (the Mouselab MDP). The goal of the benchmark test was to discover and describe strategies used in this environment by people.
\begin{table}[ht!]
    \centering
    \begin{tabular}{|l|l|}
        \hline
        \textbf{Parameter} & \textbf{Value} \\
        \hline
        exp\_id & v1.0 \\
        \hline
        num\_participants & 180 \\
        \hline
        block & train \& test \\
        \hline
        num\_clusters & 20 \\
        \hline
        num\_demos & 128 \\
        \hline
        $\mathcal{L}$ & See Section~\ref{sec:a2} \\
        \hline
        features & See Section~\ref{sec:a4} \\
        \hline
        tolerance & $1\mathrm{e}{-4}$ \\
        \hline
        change\_tolerance & $1\mathrm{e}{-5}$ \\
        \hline
        interpret\_size & 5 \\
        \hline
        ai\_tolerance & 0.025 \\
        \hline
        num\_rollouts & 10000 \\
        \hline
        num\_ai\_clusters & 4 \\
        \hline
        expert\_reward & 39.97 \\
        \hline
        max\_divergence & 0.2 \\
        \hline
        threshold & 0.5 \\
        \hline
        allowed\_predicates & See Section~\ref{sec:a3} \\
        \hline
        redundant\_predicates & See Section~\ref{sec:a3} \\
        \hline
    \end{tabular}
    \caption{Values for the parameters of Human-Interpret that were used in the benchmark test.}
    \label{tab:params2}
\end{table}

\section{Defining the Domain Specific Language}\label{sec:a2}
Our Domain Specific Language consisted of a number of logical predicates. Every predicate we defined accepts (at least) two arguments: (belief) state of the environment $b$ and computation/action $c$. In our case, the state relates to the list of expected values of nodes in the Mouselab MDP, whereas the computation is the number of the node to click, with $0$ reserved for termination. The Mouselab MDP we used in the benchmark test had the form of a tree, hence a lot of the predicates made use of notions used for the tree graph structures. The meaning of the predicates, presented below in alphabetical order, is the following:

\subsection*{A}
\paragraph*{\texttt{all}(b,c,pred$_1$,pred$_2$)}\ : All the nodes in the MDP that satisfy $pred_1$ also satisfy $pred_2$.
\paragraph*{\texttt{among}(b,c,pred$_1$,pred$_2$)}\ :  This node is among all the nodes in the MDP that satisfy $pred_1$ and inside that set it also satisfies $pred_2$.
\paragraph*{\texttt{are\_branch\_leaves\_observed}(b,c)}\ :  This node has successor leaves which are all observed.
\paragraph*{\texttt{are\_leaves\_observed}(b,c)}\ :  All leaf nodes have been observed.
\paragraph*{\texttt{are\_roots\_observed}(b,c)}\ :  All nodes on level 1 have been observed.

\subsection*{D}
\paragraph*{\texttt{depth}(b,c,d)}\ :  This node lies on level $d$.

\subsection*{H}
\paragraph*{\texttt{has\_best\_path}(b,c,list)}\ :  This node lies on a path for which the sum of expected rewards is the highest for the paths on which other nodes in $list$ lie.
\paragraph*{\texttt{has\_child\_highest\_value}(b,c,list)}\ :  This node has a child with an observed value that is higher than any other observed child's value for the nodes from $list$.
\paragraph*{\texttt{has\_child\_highest\_level\_value}(b,c)}\ :  This node's child has the maximum possible value on its level.
\paragraph*{\texttt{has\_child\_lowest\_value}(b,c,list)}\ :  This node has a child with an observed value that is lower than any other observed child's value for the nodes from $list$.
\paragraph*{\texttt{has\_child\_lowest\_level\_value}(b,c)}\ :  This node's child has the minimum posible value of its level.
\paragraph*{\texttt{has\_largest\_depth}(b,c,list)}\ :  This node is the deepest in the tree among the nodes from $list$.
\paragraph*{\texttt{has\_leaf\_highest\_value}(b,c,list)}\ :  This node has a successor that is a leaf with an observed value that is higher than any other observed successor-leaf's value for the nodes from $list$.
\paragraph*{\texttt{has\_leaf\_highest\_level\_value}(b,c)}\ :  This node leads to an uncovered leaf that has the maximum possible value on its level.
\paragraph*{\texttt{has\_leaf\_lowest\_value}(b,c,list)}\ :  This node has a successor that is a leaf with an observed value that is lower than any other observed successor-leaf's value for the nodes from $list$.
\paragraph*{\texttt{has\_leaf\_lowest\_level\_value}(b,c)}\ :  This node leads to an uncovered leaf that has the minimum possible value on its level
\paragraph*{\texttt{has\_most\_branches}(b,c,list)}\ :  This node belongs to the largest number of paths among the nodes in $list$.
\paragraph*{\texttt{has\_parent\_highest\_value}(b,c,list)}\ :  This node has a parent with an observed value that is higher than any other observed parent's value for the nodes from $list$.
\paragraph*{\texttt{has\_parent\_highest\_level\_value}(b,c)}\ :  This node's parent has the maximum possible value on its level.
\paragraph*{\texttt{has\_parent\_lowest\_value}(b,c,list)}\ :  This node has a parent with an observed value that is lower than any other observed parent's value for the nodes from $list$.
\paragraph*{\texttt{has\_parent\_lowest\_level\_value}(b,c)}\ :  This node's parent has the minimum possible value of its level.
\paragraph*{\texttt{has\_root\_highest\_value}(b,c,list)}\ :  This node has an ancestor on level 1 with an observed value that is higher than any other observed 1st-level ancestor's value for the nodes from $list$.
\paragraph*{\texttt{has\_root\_highest\_level\_value}(b,c)}\ :  This node can be accessed through an observed node on level 1 which has the highest value on level 1.
\paragraph*{\texttt{has\_root\_lowest\_value}(b,c,list)}\ :  This node has an ancestor on level 1 with an observed value that is lower than any other observed 1st-level ancestor's value for the nodes from $list$.
\paragraph*{\texttt{has\_root\_lowest\_level\_value}(b,c)}\ :  This node can be accessed through an observed node on level 1 which has the minimum value on level 1.
\paragraph*{\texttt{has\_smallest\_depth}(b,c,list)}\ :  This node is the shallowest in the tree among the nodes from $list$.

\subsection*{I}
\paragraph*{\texttt{is\_ancestor\_max\_val}(b,c)}\ :  One of the ancestors of this node is uncovered and has the maximum possible value in the MDP.
\paragraph*{\texttt{is\_leaf}(b,c)}\ :  This node is a leaf.
\paragraph*{\texttt{is\_max\_in\_branch}(b,c)}\ :  This node lies on a path with an uncovered maximum possible value in the MDP.
\paragraph*{\texttt{is\_2max\_in\_branch}(b,c)}\ :  This node lies on a path with 2 uncovered maximum possible values in the MDP.
\paragraph*{\texttt{is\_observed}(b,c)}\ :  This node was already clicked and is observed.
\paragraph*{\texttt{is\_on\_highest\_expected\_value\_path}(b,c)}\ :  This node lies on a path that has the highest expected value.
\paragraph*{\texttt{is\_positive\_observed}(b,c)}\ :  There is a node with a positive value observed.
\paragraph*{\texttt{is\_previous\_observed\_max}(b,c)}\ :  The previously observed node uncovered the maximum possible value in the MDP.
\paragraph*{\texttt{is\_previous\_observed\_max\_leaf}(b,c)}\ :  The previously observed node is a leaf and it uncovered the maximum possible value in the MDP.
\paragraph*{\texttt{is\_previous\_observed\_max\_level}(b,c)}\ :  The previously observed node uncovered a maximum possible value on that level.
\paragraph*{\texttt{is\_previous\_observed\_max\_nonleaf}(b,c)}\ :  The previously observed node isn't a leaf and it uncovered the maximum possible value in the MDP.
\paragraph*{\texttt{is\_previous\_observed\_max\_root}(b,c)}\ :  The previously observed node lies on level 1 and it uncovered the maximum possible value in the MDP.
\paragraph*{\texttt{is\_previous\_observed\_min}(b,c)}\ :  The previously observed node uncovered the minimum possible value in the MDP.
\paragraph*{\texttt{is\_previous\_observed\_min\_level}(b,c)}\ :  The previously observed node uncovered a minimum possible value on that level.
\paragraph*{\texttt{is\_previous\_observed\_parent}(b,c)}\ :  The previously observed node is the parent of this node.
\paragraph*{\texttt{is\_previous\_observed\_sibling}(b,c)}\ :  The previously observed node is one of the siblings of this node.
\paragraph*{\texttt{is\_root}(b,c)}\ :  This node is one of the nodes on level 1.
\paragraph*{\texttt{is\_successor\_max\_val}(b,c)}\ :  One of the successors of this node is uncovered and has the maximum possible value in the MDP.

\subsection*{O}
\paragraph*{\texttt{observed\_count}(b,c,n)}\ :  There are at least $n$ observed nodes.

\subsection*{T}
\paragraph*{\texttt{termination\_return}(b,c,e)}\ :  The expected reward after stopping now is $\geq e$.\newline

The DSL we used for studying Mouselab MDP policies was generated through a probabilistic context-free grammar with the following format:

 \lstset{
  basicstyle=\itshape,
  escapeinside={(*@}{@*)},
  xleftmargin=3em,
  literate={->}{$\rightarrow$}{2}
           {α}{$\alpha$}{1}
           {δ}{$\delta$}{1}
}
           
\begin{lstlisting}[mathescape=True, caption={Probabilistic context-free grammar that generates the predicates used by AI-Interpret and in consequence by Human-Interpret. Probability of each production is uniform with respect to the non-terminal symbol on its left hand-side.}]
(*@\vspace{0.5mm}@*)
START -> (*@\texttt{all}@*)(PREDS_AMONG_PRED_DEPTH)
START -> (*@\texttt{all}@*)(PREDS_AMONG_PRED_LEAF)
START -> (*@\texttt{all}@*)(PREDS_AMONG_PRED_ROOT)
START -> (*@\texttt{all}@*)(PREDS_AMONG_PRED_PARENT)
START -> (*@\texttt{all}@*)(PREDS_AMONG_PRED_CHILD)
START -> (*@\texttt{all}@*)(PREDS_AMONG_PRED_PRED)
START -> (*@\texttt{among}@*)(PREDS_AMONG_PRED_DEPTH)
START -> (*@\texttt{among}@*)(PREDS_AMONG_PRED_LEAF)
START -> (*@\texttt{among}@*)(PREDS_AMONG_PRED_ROOT)
START -> (*@\texttt{among}@*)(PREDS_AMONG_PRED_PARENT)
START -> (*@\texttt{among}@*)(PREDS_AMONG_PRED_CHILD)
START -> (*@\texttt{among}@*)(PREDS_AMONG_PRED_PRED)
START -> (*@\texttt{among}@*)(PREDS_DEPTH)
START -> (*@\texttt{among}@*)(PREDS_LEAF)
START -> (*@\texttt{among}@*)(PREDS_ROOT)
START -> (*@\texttt{among}@*)(PREDS_PARENT)
START -> (*@\texttt{among}@*)(PREDS_CHILD)
START -> (*@\texttt{among}@*)(PREDS_PRED)
START -> PRED
START -> GENERAL_PRED

PREDS_AMONG_PRED_DEPTH -> PREDS_DEPTH, AMONG_CHILD
PREDS_AMONG_PRED_DEPTH -> PREDS_DEPTH, AMONG_PARENT
PREDS_AMONG_PRED_DEPTH -> PREDS_DEPTH, AMONG_LEAF
PREDS_AMONG_PRED_DEPTH -> PREDS_DEPTH, AMONG_ROOT

PREDS_AMONG_PRED_LEAF -> PREDS_LEAF, AMONG_CHILD
PREDS_AMONG_PRED_LEAF -> PREDS_LEAF, AMONG_PARENT
PREDS_AMONG_PRED_LEAF -> PREDS_LEAF, AMONG_ROOT
PREDS_AMONG_PRED_LEAF -> PREDS_LEAF, AMONG_PRED

PREDS_AMONG_PRED_ROOT -> PREDS_ROOT, AMONG_CHILD
PREDS_AMONG_PRED_ROOT -> PREDS_ROOT, AMONG_PARENT
PREDS_AMONG_PRED_ROOT -> PREDS_ROOT, AMONG_LEAF
PREDS_AMONG_PRED_ROOT -> PREDS_ROOT, AMONG_PRED

PREDS_AMONG_PRED_PARENT -> PREDS_PARENT, AMONG_CHILD
PREDS_AMONG_PRED_PARENT -> PREDS_PARENT, AMONG_LEAF
PREDS_AMONG_PRED_PARENT -> PREDS_PARENT, AMONG_ROOT
PREDS_AMONG_PRED_PARENT -> PREDS_PARENT, AMONG_PRED

PREDS_AMONG_PRED_CHILD -> PREDS_CHILD, AMONG_PARENT
PREDS_AMONG_PRED_CHILD -> PREDS_CHILD, AMONG_LEAF
PREDS_AMONG_PRED_CHILD -> PREDS_CHILD, AMONG_ROOT
PREDS_AMONG_PRED_CHILD -> PREDS_CHILD, AMONG_PRED

PREDS_AMONG_PRED_PRED -> PREDS_PRED, AMONG_CHILD
PREDS_AMONG_PRED_PRED -> PREDS_PRED, AMONG_PARENT
PREDS_AMONG_PRED_PRED -> PREDS_PRED, AMONG_LEAF
PREDS_AMONG_PRED_PRED -> PREDS_PRED, AMONG_ROOT
PREDS_AMONG_PRED_PRED -> PREDS_PRED, AMONG_PRED

PREDS_DEPTH -> DEPTH
PREDS_DEPTH -> DEPTH (*@\texttt{and}@*) LEAF
PREDS_DEPTH -> DEPTH (*@\texttt{and}@*) ROOT
PREDS_DEPTH -> DEPTH (*@\texttt{and}@*) PARENT
PREDS_DEPTH -> DEPTH (*@\texttt{and}@*) CHILD
PREDS_DEPTH -> DEPTH (*@\texttt{and}@*) DEPTH
PREDS_DEPTH -> DEPTH (*@\texttt{and}@*) PRED

PREDS_LEAF -> LEAF
PREDS_LEAF -> LEAF (*@\texttt{and}@*) LEAF
PREDS_LEAF -> LEAF (*@\texttt{and}@*) ROOT
PREDS_LEAF -> LEAF (*@\texttt{and}@*) PARENT
PREDS_LEAF -> LEAF (*@\texttt{and}@*) CHILD
PREDS_LEAF -> LEAF (*@\texttt{and}@*) PRED

PREDS_ROOT -> ROOT
PREDS_ROOT -> ROOT (*@\texttt{and}@*) ROOT
PREDS_ROOT -> ROOT (*@\texttt{and}@*) PARENT
PREDS_ROOT -> ROOT (*@\texttt{and}@*) CHILD
PREDS_ROOT -> ROOT (*@\texttt{and}@*) PRED

PREDS_PARENT -> PARENT
PREDS_PARENT -> PARENT (*@\texttt{and}@*) PARENT
PREDS_PARENT -> PARENT (*@\texttt{and}@*) CHILD
PREDS_PARENT -> PARENT (*@\texttt{and}@*) PRED

PREDS_PARENT -> PARENT
PREDS_PARENT -> PARENT (*@\texttt{and}@*) PARENT
PREDS_PARENT -> PARENT (*@\texttt{and}@*) CHILD
PREDS_PARENT -> PARENT (*@\texttt{and}@*) PRED

PREDS_CHILD -> CHILD
PREDS_CHILD -> CHILD (*@\texttt{and}@*) CHILD
PREDS_CHILD -> CHILD (*@\texttt{and}@*) PRED

PREDS_PRED -> PRED
PREDS_PRED -> PRED (*@\texttt{and}@*) PRED

AMONG_PRED -> (*@\texttt{has\_lowest\_depth $\mid$ has\_largest\_depth}@*)
AMONG_PRED -> (*@\texttt{has\_best\_path $\mid$ has\_most\_paths}@*)

AMONG_CHILD -> (*@\texttt{has\_child\_highest\_value $\mid$  has\_child\_lowest\_value}@*)
AMONG_PARENT -> (*@\texttt{has\_parent\_highest\_value $\mid$  has\_parent\_lowest\_value}@*)
AMONG_LEAF -> (*@\texttt{has\_leaf\_highest\_value $\mid$  has\_leaf\_lowest\_value}@*)
AMONG_ROOT -> (*@\texttt{has\_root\_highest\_value $\mid$  has\_root\_lowest\_value}@*)

DEPTH -> (*@\texttt{depth(DEP) $\mid$  not(depth(DEP))}@*)
LEAF -> (*@\texttt{has\_leaf\_highest\_level\_value $\mid$  has\_leaf\_lowest\_level\_value}@*)
LEAF -> (*@\texttt{not(has\_leaf\_highest\_level\_value)}@*)
LEAF -> (*@\texttt{not(has\_leaf\_lowest\_level\_value)}@*)
ROOT -> (*@\texttt{has\_root\_highest\_level\_value $\mid$  has\_root\_lowest\_level\_value}@*)
ROOT -> (*@\texttt{not(has\_root\_highest\_level\_value)}@*)
ROOT -> (*@\texttt{not(has\_root\_lowest\_level\_value)}@*)
PARENT -> (*@\texttt{has\_parent\_highest\_level\_value $\mid$ has\_parent\_lowest\_level\_value}@*)
PARENT -> (*@\texttt{not(has\_parent\_highest\_level\_value)}@*)
PARENT -> (*@\texttt{not(has\_parent\_lowest\_level\_value)}@*)
CHILD -> (*@\texttt{has\_child\_highest\_level\_value $\mid$  has\_child\_lowest\_level\_value}@*)
CHILD -> (*@\texttt{not(has\_child\_highest\_level\_value)}@*)
CHILD -> (*@\texttt{not(has\_child\_lowest\_level\_value)}@*)
PRED -> (*@\texttt{is\_leaf $\mid$  is\_root $\mid$  is\_max\_in\_branch}@*)
PRED -> (*@\texttt{is\_2max\_in\_branch $\mid$  are\_branch\_leaves\_observed}@*)
PRED -> (*@\texttt{not(is\_leaf) $\mid$  not(is\_root) $\mid$  not(is\_max\_in\_branch)}@*)
PRED -> (*@\texttt{not(is\_2max\_in\_branch)}@*)
PRED -> (*@\texttt{not(are\_branch\_leaves\_observed) $\mid$  not(is\_observed)}@*)

GENERAL_PRED -> (*@\texttt{is\_previous\_observed\_max $\mid$  is\_positive\_observed}@*)
GENERAL_PRED -> (*@\texttt{are\_leaves\_observed}@*)
GENERAL_PRED -> (*@\texttt{are\_roots\_observed $\mid$  is\_previous\_observed\_positive}@*)
GENERAL_PRED -> (*@\texttt{is\_previous\_observed\_parent}@*)
GENERAL_PRED -> (*@\texttt{is\_previous\_observed\_sibling}@*)
GENERAL_PRED -> (*@\texttt{is\_previous\_observed\_min}@*)
GENERAL_PRED -> (*@\texttt{is\_previous\_observed\_max\_nonleaf}@*)
GENERAL_PRED -> (*@\texttt{is\_previous\_observed\_max\_leaf}@*)
GENERAL_PRED -> (*@\texttt{is\_previous\_observed\_max\_root}@*)
GENERAL_PRED -> (*@\texttt{is\_previous\_observed\_max\_level(DEP)}@*)
GENERAL_PRED -> (*@\texttt{is\_previous\_observed\_min\_level(DEP)}@*)
GENERAL_PRED -> (*@\texttt{observed\_count(NUM) $\mid$  termination\_return(RET)}@*)

NUM -> (*@\texttt{1 $\mid$  2 $\mid$  3 $\mid$  4 $\mid$  5 $\mid$  6 $\mid$  7 $\mid$  8}@*)
DEP -> (*@\texttt{1 $\mid$  2 $\mid$  3}@*)
RET -> (*@\texttt{-30 $\mid$  -25 $\mid$  -15 $\mid$  -10 $\mid$  0 $\mid$  10 $\mid$  15 $\mid$  25 $\mid$  30}@*)
\end{lstlisting}

\section{Defining redundant and allowed predicates}\label{sec:a3}
The \texttt{redundant\_predicates} parameter of Human-Interpret controls which predicates are to be removed from the DNF formula before it is turned into linear temporal logic formula. The \texttt{allowed\_predicates} parameter specifies which predicates are considered for the until and unless conditions. We used the following values:

\begin{align*}
\texttt{allowed\_predicates} = &\ [\text{All predicates derived from GENERAL\_PRED in the CFG above}, \\
                               & \ is\_max\_in\_branch,\  are\_branch\_leaves\_observed] \\ 
                               \\
\texttt{redundant\_predicates} = &\ [\text{All predicates of the type \texttt{all}(b,c,pred$_1$,pred$_2$)}, \\
                               & \ \texttt{allowed\_predicates}]                               
\end{align*}

\section{Defining features for the softmax models}\label{sec:a4}
The generative models created in the generative step of the Human-Interpret algorithm take form of softmax functions shown in Equation~\ref{eq:softmax}. Those function are defined on a set of features derived from the Mouselab MDP in a given (belief) state $b$ while taking a given computation/action $c$ (node to click). We used the following features in the benchmark test:

\subsection*{C}
\paragraph*{\texttt{count_observed_node_branch}(b,c)}\ :  What is the minimum of the number of observed nodes on branches that pass through the given node.

\subsection*{D}
\paragraph*{\texttt{depth_count}(b,c)}\ :  What is the number of observed nodes on the same depth as the given node.
\paragraph*{\texttt{depth}(b,c)}\ :  What is the depth of the given node.

\subsection*{G}
\paragraph*{\texttt{get_level_observed_std}(b,c)}\ :  What is the standard deviation for the values of observed nodes at the same level as the given node.

\subsection*{H}
\paragraph*{\texttt{hp_0}(b,c)}\ :  Does the path that the node lies on has a value greater than 0.

\subsection*{I}
\paragraph*{\texttt{immediate_successor_count}(b,c)}\ :  What is the number of observed children of the given node.
\paragraph*{\texttt{is_leaf}(b,c)}\ :  Is the given node a leaf or not.
\paragraph*{\texttt{is_previous_max}(b,c)}\ :  Did the previously observed node uncover the maximum possible value in the MDP or not. 
\paragraph*{\texttt{is_root}(b,c)}\ :  Is the given node a root or not.

\subsection*{M}
\paragraph*{\texttt{most_promising}(b,c)}\ :  Whether the given node lies on the path with the highest expected reward or not.

\subsection*{O}
\paragraph*{\texttt{observed_height}(b,c)}\ :  What is the maximum number of consecutively observed nodes on the path of the given node, starting from its children.

\subsection*{P}
\paragraph*{\texttt{parent_observed}(b,c)}\ :  Whether the parent of the given node is observed or not.
\paragraph*{\texttt{previous_observed_successor}(b,c)}\ :  Whether the previously observed node was the parent of the given node or not.

\subsection*{S}
\paragraph*{\texttt{second_most_promising}(b,c)}\ :  Whether the given node lies on the path with the second highest expected reward or not.
\paragraph*{\texttt{siblings_count}(b,c)}\ :  What is the number of observed siblings of the given node.
\paragraph*{\texttt{soft_satisficing}(b,c)}\ :  What is the expected reward of terminating now and traversing the most promising path.
\paragraph*{\texttt{successor_uncertainty}(b,c)}\ :  What is the total standard deviation for the children of the given node.

\subsection*{T}
\paragraph*{\texttt{termination_constant}(b,c)}\ :  The value of this feature is 0 for the termination action and is -1 for all the nodes. 

\subsection*{U}
\paragraph*{\texttt{uncertainty}(b,c)}\ :  What is the total standard deviation for the nodes on the same depth as the given node.

\newpage
\section{Benchmark test results}
As mentioned in the main text, the 21 runs of Human-Interpret returned a set of 21 unique strategies. Here, we extend Table 3 from the main text, and in Table~\ref{tab:fullstats} present all of the discovered strategies alongside their LTL formulas. Note that sometimes strategies discovered in different runs differed by 1 predicate (e.g. \texttt{depth(3)} vs. \texttt{is\_leaf}) and so we listed equivalent predicates in brackets.

\begin{longtable}[H]{p{0.03\textwidth}p{0.51\textwidth}p{0.06\textwidth}p{0.03\textwidth}p{0.03\textwidth}p{0.03\textwidth}p{0.02\textwidth}p{0.02\textwidth}} \toprule
\multirow{2}*{ID} & \multirow{2}*{Generated formulas and descriptions}& \multicolumn{6}{c}{Statistics} \\
\cmidrule{3-8}
& & FR & FCF & FON & FPO & N & GT\\
\midrule
\endhead

\multirow{23}*{1} & \multicolumn{6}{c}{\textbf{Summary: }Search for the best possible outcome sibling by sibling.} \\
\cmidrule{2-8}
& \multirow{1}{0.5\textwidth}{\textbf{Description: }
\begin{enumerate}
    \item Click on a node satisfying all of the following conditions: \begin{enumerate}[label=-]
        \item it is an unobserved leaf.
    \end{enumerate} 
    \item Unless the previously observed node uncovers a 48, in which case stop at the previous step, click on a node satisfying all of the following conditions: 
    \begin{enumerate}[label=-]
        \item it is an unobserved leaf.
    \end{enumerate}
    click in this way under the condition that:
    \begin{enumerate}[label=-]
        \item the previously observed node was its sibling.
    \end{enumerate}
    \item GOTO step 1 unless all the leaves are observed or the previously observed node uncovers a 48.
\end{enumerate}
\textbf{LTL formula: }\texttt{among(not(is\_observed) and is\_leaf) AND NEXT among(not(is\_observed) and is\_leaf) and is\_previous\_observed\_sibling UNLESS is\_previous\_observed\_max\_leaf 
\\
LOOP FROM among(not(is\_observed) and is\_leaf) UNLESS (are\_leaves\_observed or is\_previous\_observed\_max\_leaf)}} 
& 32.3\% & 0.98 & 0.91 & 0.84 & 21 & Y\\
& & & & & & & \\
& & & & & & & \\
& & & & & & & \\
& & & & & & & \\
& & & & & & & \\
& & & & & & & \\
& & & & & & & \\
& & & & & & & \\
& & & & & & & \\
& & & & & & & \\
& & & & & & & \\
& & & & & & & \\
& & & & & & & \\
& & & & & & & \\
& & & & & & & \\
& & & & & & & \\
& & & & & & & \\
& & & & & & & \\
& & & & & & & \\
& & & & & & & \\
& & & & & & & \\
\cmidrule{1-8}

\multirow{3}*{2} & \multicolumn{6}{c}{\textbf{Summary: }No planning.} \\
\cmidrule{2-8}
& \textbf{Description: }Do not click. 
& 24.6\% & 1 & 0.51 & 0.41 & 20 & Y\\
& \textbf{LTL formula: }\texttt{False UNTIL IT STOPS APPLYING} & & & & & & \\
\cmidrule{1-8}

\multirow{8}*{3} & \multicolumn{6}{c}{\textbf{Summary: }Click all immediate outcomes.} \\
\cmidrule{2-8}
& \multirow{1}{0.5\textwidth}{\textbf{Description: }
\begin{enumerate}
    \item Click on the nodes satisfying all of the following conditions: \begin{enumerate}[label=-]
        \item they are unobserved roots. 
    \end{enumerate} 
    Repeat this step as long as possible.
\end{enumerate}
\textbf{LTL formula: }\texttt{among(not(is\_observed) and is\_root) UNTIL IT STOPS APPLYING}}
& 4.2\% & 1 & 0.74 & 0.61 & 3 & Y\\
& & & & & & & \\
& & & & & & & \\
& & & & & & & \\
& & & & & & & \\
& & & & & & & \\
& & & & & & & \\
\cmidrule{1-8}

\multirow{11}*{4} & \multicolumn{6}{c}{\textbf{Summary: }Search for the best possible outcome.} \\
\cmidrule{2-8}
& \multirow{4}{0.5\textwidth}{\textbf{Description: }
\begin{enumerate}
    \item Click on the nodes satisfying all of the following conditions: \begin{enumerate}[label=-]
        \item they are unobserved leaves. 
    \end{enumerate} 
    Repeat this step until all the leaves are observed or the previously observed node uncovers a 48.
\end{enumerate}
\textbf{LTL formula: }\texttt{among(not(is\_observed) and is\_leaf) UNTIL (are\_leaves\_observed or is\_previous\_observed\_max\_leaf)}} 
& 17.2\% & 0.88 & 0.87 & 0.79 & 21 & Y\\
& & & & & & & \\
& & & & & & & \\
& & & & & & & \\
& & & & & & & \\
& & & & & & & \\
& & & & & & & \\
& & & & & & & \\
& & & & & & & \\
& & & & & & & \\
\cmidrule{1-8}

\multirow{58}*{6} & \multicolumn{6}{c}{\textbf{Summary: }Click all final outcomes sibling by sibling.} \\
\cmidrule{2-8}
& \multirow{1}{0.5\textwidth}{\textbf{Description: }
\begin{enumerate}
    \item Click on a node satisfying all of the following conditions: \begin{enumerate}[label=-]
        \item it is an unobserved leaf.
    \end{enumerate} 
    \item Click on a node satisfying all of the following conditions: 
    \begin{enumerate}[label=-]
        \item it is an unobserved leaf.
        \item the previously observed node was its sibling 
    \end{enumerate}
    \item GOTO step 1.
\end{enumerate}
\textbf{LTL formula: }\texttt{among(not(is\_observed) and is\_leaf) AND NEXT among(not(is\_observed) and is\_leaf) and is\_previous\_observed\_sibling
\\
LOOP FROM among(not(is\_observed) and is\_leaf)}
\\
\textbf{Description: }
\begin{enumerate}
    \item Click on the nodes satisfying all of the following conditions: 
    \begin{enumerate}[label=-]
        \item they are unobserved nodes that lead to leaves whose value is different from -48
        \item they are located on the highest level considering the unobserved nodes that lead to leaves whose value is different from -48.
    \end{enumerate}
    Repeat this step until a node with a positive value is observed or the previously observed node uncovers a -48.
    \item Click on a node satisfying all of the following conditions: 
    \begin{enumerate}[label=-]
        \item it is an unobserved node that leads to a leaf whose value is different from -48
        \item it is located on the highest level considering the unobserved nodes that lead to leaves whose value is different from -48
        \item it is the previously observed node was its sibling. 
    \end{enumerate}
    \item GOTO step 1 unless all the leaves are observed or all the roots are observed.
\end{enumerate}
\textbf{LTL formula: }\texttt{among(not(has\_leaf\_lowest\_level\_value) and not(is\_observed) : has\_largest\_depth) UNTIL (is\_positive\_observed or is\_previous\_observed\_min\_level(3)) AND NEXT among(not(has\_leaf\_lowest\_level\_value) and not(is\_observed) : has\_largest\_depth) and is\_previous\_observed\_sibling 
\\
LOOP FROM among(not(has\_leaf\_lowest\_level\_value) and not(is\_observed) : has\_largest\_depth) UNLESS (are\_leaves\_observed or are\_roots\_observed)}} 
& 33.5\% & 0.74 & 0.65 & 0.47 & 18 & N\\
& & & & & & & \\
& & & & & & & \\
& & & & & & & \\
& & & & & & & \\
& & & & & & & \\
& & & & & & & \\
& & & & & & & \\
& & & & & & & \\
& & & & & & & \\
& & & & & & & \\
& & & & & & & \\
& & & & & & & \\
& & & & & & & \\
& & & & & & & \\
& & & & & & & \\
& & & & & & & \\
& & & & & & & \\
& & & & & & & \\
& & & & & & & \\
& & & & & & & \\
& & & & & & & \\
& & & & & & & \\
& & & & & & & \\
& & & & & & & \\
& & & & & & & \\
& & & & & & & \\
& & & & & & & \\
& & & & & & & \\
& & & & & & & \\
& & & & & & & \\
& & & & & & & \\
& & & & & & & \\
& & & & & & & \\
& & & & & & & \\
& & & & & & & \\
& & & & & & & \\
& & & & & & & \\
& & & & & & & \\
& & & & & & & \\
& & & & & & & \\
& & & & & & & \\
& & & & & & & \\
& & & & & & & \\
& & & & & & & \\
& & & & & & & \\
& & & & & & & \\
& & & & & & & \\
& & & & & & & \\
& & & & & & & \\
& & & & & & & \\
& & & & & & & \\
& & & & & & & \\
\cmidrule{1-8}

\pagebreak

\multirow{12}*{5} & \multicolumn{6}{c}{\textbf{Summary: }Best first search.} \\
\cmidrule{2-8}
& \multirow{9}{0.55\textwidth}{\textbf{Description: }
\begin{enumerate} 
    \item Click on the nodes satisfying all of the following conditions: \begin{enumerate}[label=-] 
        \item they are unobserved nodes 
        \item they have parents with the highest values considering the parents of other unobserved nodes.
    \end{enumerate} 
    Repeat this step as long as possible. 
\end{enumerate}
\textbf{LTL formula: }\texttt{among(not(is\_observed) : has\_parent\_highest\_value) UNTIL IT STOPS APPLYING}} 
& 3.8\% & 0.63 & 0.67 & 0.64 & 7 & Y\\
& & & & & & & \\
& & & & & & & \\
& & & & & & & \\
& & & & & & & \\
& & & & & & & \\
& & & & & & & \\
& & & & & & & \\
& & & & & & & \\
& & & & & & & \\
& & & & & & & \\
\cmidrule{1-8}

\multirow{20}*{7} & \multicolumn{6}{c}{\textbf{Summary: }Depth first search.} \\
\cmidrule{2-8}
& \multirow{9}{0.55\textwidth}{\textbf{Description: }
\begin{enumerate}
    \item Click on a node satisfying all of the following conditions: \begin{enumerate}[label=-]
        \item it is an unobserved node
        \item it has a parent with the lowest value considering the parents of other unobserved nodes.
    \end{enumerate}
    \item Click on the nodes satisfying all of the following conditions: \begin{enumerate}[label=-]
        \item they are unobserved non-roots
        \item they have parents with the highest values considering the parents of other unobserved non-roots.
    \end{enumerate}
    Repeat this step as long as possible.
    \item GOTO step 1.
\end{enumerate}
\textbf{LTL formula: }\texttt{among(not(is\_observed) : has\_parent\_[lowest|highest]\_value) AND NEXT among(not(is\_observed) and not(is\_root) : has\_parent\_highest\_value) UNTIL IT STOPS APPLYING 
\\
LOOP FROM among(not(is\_observed) : has\_parent\_[lowest|highest]\_value)}} 
& 9.1\% & 0.84 & 0.72 & 0.6 & 7 & Y\\
& & & & & & & \\
& & & & & & & \\
& & & & & & & \\
& & & & & & & \\
& & & & & & & \\
& & & & & & & \\
& & & & & & & \\
& & & & & & & \\
& & & & & & & \\
& & & & & & & \\
& & & & & & & \\
& & & & & & & \\
& & & & & & & \\
& & & & & & & \\
& & & & & & & \\
& & & & & & & \\
& & & & & & & \\
& & & & & & & \\
& & & & & & & \\
& & & & & & & \\
& & & & & & & \\
\cmidrule{1-8}

\multirow{13}*{9} & \multicolumn{6}{c}{\textbf{Summary: }Click on a node and all its successors. Repeat or terminate.} \\
\cmidrule{2-8}
& \multirow{9}{0.55\textwidth}{\textbf{Description: }\begin{enumerate} \item Click on a random node or terminate. \item Click on the nodes satisfying all of the following: \begin{enumerate}[label=-] \item they are unobserved non-roots \item they have parents with the lowest (highest) values considering the parents of other unobserved non-roots.\end{enumerate} Repeat this step as long as possible. \item GOTO step 1.\end{enumerate}\textbf{LTL formula: }\texttt{True AND NEXT among(not(is\_observed) and not(is\_root) : has\_parent\_[lowest|highest]\_value) UNTIL IT STOPS APPLYING GO TO True}} 
& 4\% & 0.58 & 0.84 & 0.73 & 5 & Y\\
& & & & & & & \\
& & & & & & & \\
& & & & & & & \\
& & & & & & & \\
& & & & & & & \\
& & & & & & & \\
& & & & & & & \\
& & & & & & & \\
& & & & & & & \\
& & & & & & & \\
& & & & & & & \\
& & & & & & & \\
\cmidrule{1-8}

\pagebreak

\multirow{28}*{8} & \multicolumn{6}{c}{\textbf{Summary: }Depth first search until finding the best final outcome.} \\
\cmidrule{2-8}
& \multirow{9}{0.55\textwidth}{\textbf{Description: }
\begin{enumerate}
    \item Click on a node satisfying all of the following conditions: \begin{enumerate}[label=-]
        \item it is an unobserved node
        \item it has a parent with the lowest value considering the parents of other unobserved nodes.
    \end{enumerate}
    Click in this way under the condition that:
    \begin{enumerate}[label=-]
        \item the previously observed node uncovered something else than a 48.
    \end{enumerate}
    \item Click on the nodes satisfying all of the following conditions: \begin{enumerate}[label=-]
        \item they are unobserved non-roots
        \item they have parents with the highest values considering the parents of other unobserved non-roots.
    \end{enumerate}
    Repeat this step as long as possible.
    \item GOTO step 1 unless all the leaves are observed or the previously observed node uncovers a 48.
\end{enumerate}
\textbf{LTL formula: }\texttt{among(not(is\_observed) : has\_parent\_[lowest|highest]]\_value) and not(is\_previous\_observed\_max\_leaf) AND NEXT among(not(is\_observed) and not(is\_root) : has\_parent\_highest\_value) UNTIL IT STOPS APPLYING 
\\
LOOP FROM among(not(is\_observed) : has\_parent\_[lowest|highest]\_value) and not(is\_previous\_observed\_max\_leaf)}} 
& 6.0\% & 0.92 & 0.71 & 0.71 & 1 & Y\\
& & & & & & & \\
& & & & & & & \\
& & & & & & & \\
& & & & & & & \\
& & & & & & & \\
& & & & & & & \\
& & & & & & & \\
& & & & & & & \\
& & & & & & & \\
& & & & & & & \\
& & & & & & & \\
& & & & & & & \\
& & & & & & & \\
& & & & & & & \\
& & & & & & & \\
& & & & & & & \\
& & & & & & & \\
& & & & & & & \\
& & & & & & & \\
& & & & & & & \\
& & & & & & & \\
& & & & & & & \\
& & & & & & & \\
& & & & & & & \\
& & & & & & & \\
& & & & & & & \\
& & & & & & & \\
\cmidrule{1-8}

\multirow{20}*{11} & \multicolumn{6}{c}{\textbf{Summary: }Depth first search on a random branch.} \\
\cmidrule{2-8}
& \multirow{9}{0.55\textwidth}{\textbf{Description: }
\begin{enumerate}
    \item Click on a node satisfying all of the following conditions: \begin{enumerate}[label=-]
        \item it is an unobserved node
        \item it has a parent with the lowest value considering the parents of other unobserved nodes.
    \end{enumerate}
    \item Click on the nodes satisfying all of the following conditions: \begin{enumerate}[label=-]
        \item they are unobserved non-roots
        \item they have parents with the highest values considering the parents of other unobserved non-roots.
    \end{enumerate}
    Repeat this step as long as possible.
\end{enumerate}
\textbf{LTL formula: }\texttt{among(not(has\_child\_lowest\_level\_value) and not(is\_observed) : has\_parent\_[lowest|highest]\_value) AND NEXT among(not(is\_observed) and not(is\_root) : has\_parent\_highest\_value) UNTIL IT STOPS APPLYING}} 
& 3.8\% & 0.75 & 0.63 & 0.63 & 1 & N\\
& & & & & & & \\
& & & & & & & \\
& & & & & & & \\
& & & & & & & \\
& & & & & & & \\
& & & & & & & \\
& & & & & & & \\
& & & & & & & \\
& & & & & & & \\
& & & & & & & \\
& & & & & & & \\
& & & & & & & \\
& & & & & & & \\
& & & & & & & \\
& & & & & & & \\
& & & & & & & \\
& & & & & & & \\
& & & & & & & \\
& & & & & & & \\
& & & & & & & \\
\cmidrule{1-8}

\pagebreak

\multirow{12}*{14} & \multicolumn{6}{c}{\textbf{Summary: }Click all immediate outcomes and continue at random.} \\
\cmidrule{2-8}
& \multirow{1}{0.5\textwidth}{\textbf{Description: }
\begin{enumerate}
    \item Click on the nodes satisfying all of the following conditions: \begin{enumerate}[label=-]
        \item they are unobserved roots. 
    \end{enumerate} 
    Repeat this step until all the roots are observed.
    \item Terminate or click on some random nodes and then terminate. Repeat this step as long as possible.
\end{enumerate}
\textbf{LTL formula: }\texttt{among(not(is\_observed) and is\_root) UNTIL are\_roots\_observed AND NEXT True UNTIL IT STOPS APPLYING}}
& 2\% & 0.68 & 0.99 & 0.98 & 4 & N\\
& & & & & & & \\
& & & & & & & \\
& & & & & & & \\
& & & & & & & \\
& & & & & & & \\
& & & & & & & \\
& & & & & & & \\
& & & & & & & \\
& & & & & & & \\
& & & & & & & \\
& & & & & & & \\
\cmidrule{1-8}

\multirow{30}*{16} & \multicolumn{6}{c}{\textbf{Summary: }Click a random path or a branch.} \\
\cmidrule{2-8}
& \multirow{1}{0.5\textwidth}{\textbf{Description: }
\begin{enumerate}
    \item Click on a node satisfying all of the following conditions: \begin{enumerate}[label=-]
        \item it is a leaf
        \item lies on a best path. 
    \end{enumerate} 
    \item Click on the nodes satisfying all of the following conditions: \begin{enumerate}[label=-]
        \item they are leaves
        \item lie on best paths. 
    \end{enumerate}
    Click in this way as long as:
    \begin{enumerate}[label=-]
        \item the previously observed node was their sibling.
    \end{enumerate}
    Repeat this step until a node with a positive value is observed.
    \item Click on the nodes satisfying all of the following conditions: \begin{enumerate}[label=-]
        \item they are unobserved nodes
        \item they have children with the highest values considering the children of other unobserved nodes. 
    \end{enumerate}
    Repeat this step as long as possible.
\end{enumerate}
\textbf{LTL formula: }\texttt{among(is\_leaf : has\_best\_path) AND NEXT among(is\_leaf : has\_best\_path) and is\_previous\_observed\_sibling UNTIL is\_positive\_observed AND NEXT among(not(is\_observed) : has\_child\_highest\_value) UNTIL IT STOPS APPLYING}}
& 5.7\% & 0.55 & 0.47 & 0.34 & 3 & N\\
& & & & & & & \\
& & & & & & & \\
& & & & & & & \\
& & & & & & & \\
& & & & & & & \\
& & & & & & & \\
& & & & & & & \\
& & & & & & & \\
& & & & & & & \\
& & & & & & & \\
& & & & & & & \\
& & & & & & & \\
& & & & & & & \\
& & & & & & & \\
& & & & & & & \\
& & & & & & & \\
& & & & & & & \\
& & & & & & & \\
& & & & & & & \\
& & & & & & & \\
& & & & & & & \\
& & & & & & & \\
& & & & & & & \\
& & & & & & & \\
& & & & & & & \\
& & & & & & & \\
& & & & & & & \\
\cmidrule{1-8}

\multirow{8}*{13} & \multicolumn{6}{c}{\textbf{Summary: }Click all final outcomes.} \\
\cmidrule{2-8}
& \multirow{1}{0.5\textwidth}{\textbf{Description: }
\begin{enumerate}
    \item Click on the nodes satisfying all of the following conditions: \begin{enumerate}[label=-]
        \item they are unobserved leaves. 
    \end{enumerate} 
    Repeat this step until it stops applying.
\end{enumerate}
\textbf{LTL formula: }\texttt{among(not(is\_observed) and is\_leaf) UNTIL IT STOPS APPLYING}}
& 5.6\% & 0.69 & 0.74 & 0.55 & 6 & Y\\
& & & & & & & \\
& & & & & & & \\
& & & & & & & \\
& & & & & & & \\
& & & & & & & \\
& & & & & & & \\
\cmidrule{1-8}

\pagebreak

\multirow{45}*{20} & \multicolumn{6}{c}{\textbf{Summary: }Click two sibling final outcomes and maybe one random outcome.} \\
\cmidrule{2-8}
& \multirow{1}{0.5\textwidth}{\textbf{Description\footnote{One strategy listed steps 1 and 2 twice}: }
\begin{enumerate}
    \item Click on random nodes. Do not click on the nodes satisfying either of the following conditions:
    \begin{enumerate}[label=-]
        \item they are non-leaves that have children with the non-highest value on their level.
    \end{enumerate} 
    \item Click on random nodes. Click in this way as long as:
    \begin{enumerate}[label=-]
        \item the previously observed node was their sibling.
    \end{enumerate}
    Do not click on the nodes satisfying either of the following conditions: 
    \begin{enumerate}[label=-]
        \item they are non-leaves that have children with the non-highest value on their level.
    \end{enumerate}
    Repeat this step until this node belongs to a subtree with all leaves already observed.
    \item Click on a random node or terminate.
\end{enumerate}
\textbf{LTL formula: }\texttt{not(among(not(has\_child\_highest\_level\_value) and not(is\_leaf))) UNTIL (is\_positive\_observed or is\_previous\_observed\_min\_level(3)) AND NEXT not(among(not(has\_child\_highest\_level\_value) and not(is\_leaf))) and is\_previous\_observed\_sibling UNTIL[are\_branch\_leaves\_observed AND NEXT True}
\\
\textbf{Description: }
\begin{enumerate}
    \item Click on a node satisfying all of the following conditions:
    \begin{enumerate}[label=-]
        \item it is an unobserved leaf.
    \end{enumerate}
    \item Click on random nodes. Click in this way as long as:
    \begin{enumerate}[label=-]
        \item the previously observed node was their sibling.
    \end{enumerate}
    Do not click on the nodes satisfying either of the following conditions: 
    \begin{enumerate}[label=-]
        \item they are nodes belonging to a subtree with some unobserved leaves.
    \end{enumerate}
    Repeat this step as long as possible.
\end{enumerate}
\textbf{LTL formula: }\texttt{among(is\_leaf and not(is\_observed)) AND NEXT is\_previous\_observed\_sibling and not(are\_branch\_leaves\_observed)) UNTIL IT STOPS APPLYING}}
& 5.9\% & 0.45 & 0.48 & 0.25 & 5 & N\\
& & & & & & & \\
& & & & & & & \\
& & & & & & & \\
& & & & & & & \\
& & & & & & & \\
& & & & & & & \\
& & & & & & & \\
& & & & & & & \\
& & & & & & & \\
& & & & & & & \\
& & & & & & & \\
& & & & & & & \\
& & & & & & & \\
& & & & & & & \\
& & & & & & & \\
& & & & & & & \\
& & & & & & & \\
& & & & & & & \\
& & & & & & & \\
& & & & & & & \\
& & & & & & & \\
& & & & & & & \\
& & & & & & & \\
& & & & & & & \\
& & & & & & & \\
& & & & & & & \\
& & & & & & & \\
& & & & & & & \\
& & & & & & & \\
& & & & & & & \\
& & & & & & & \\
& & & & & & & \\
& & & & & & & \\
& & & & & & & \\
& & & & & & & \\
& & & & & & & \\
& & & & & & & \\
& & & & & & & \\
& & & & & & & \\
& & & & & & & \\
& & & & & & & \\
& & & & & & & \\
& & & & & & & \\
& & & & & & & \\
& & & & & & & \\
& & & & & & & \\
\cmidrule{2-8}

\pagebreak

\multirow{18}*{10} & \multicolumn{6}{c}{\textbf{Summary: }Search for the best final outcome in pairs.} \\
\cmidrule{2-8}
& \multirow{9}{0.5\textwidth}{\textbf{Description: }
\begin{enumerate}
    \item Click on a node satisfying all of the following conditions: \begin{enumerate}[label=-]
        \item it is an unobserved leaf
    \end{enumerate} 
    \item Click on a nodes satisfying all of the following conditions: \begin{enumerate}[label=-]
        \item it is an unobserved leaf
    \end{enumerate}
    Click in this way under the condition that
    \begin{enumerate}[label=-]
        \item the previously observed node was its sibling.
    \end{enumerate}
    \item 3. GOTO step 1 unless all the leaves are observed or the previously observed node uncovers a 48.
\end{enumerate}
\textbf{LTL formula: }\texttt{among(is\_leaf and not(is\_observed)) AND NEXT among(is\_leaf and not(is\_observed)) and is\_previous\_observed\_sibling
\\
LOOP FROM among(is\_leaf and not(is\_observed)) UNLESS (are\_leaves\_observed or [is\_previous\_observed\_max\_leaf | is\_previous\_observed\_max])
}}
& 4\% & 0.58 & 0.84 & 0.73 & 2 & N\\
& & & & & & & \\
& & & & & & & \\
& & & & & & & \\
& & & & & & & \\
& & & & & & & \\
& & & & & & & \\
& & & & & & & \\
& & & & & & & \\
& & & & & & & \\
& & & & & & & \\
& & & & & & & \\
& & & & & & & \\
& & & & & & & \\
& & & & & & & \\
& & & & & & & \\
& & & & & & & \\
& & & & & & & \\
& & & & & & & \\
& & & & & & & \\
& & & & & & & \\
& & & & & & & \\
\cmidrule{1-8}

\multirow{18}*{15} & \multicolumn{6}{c}{\textbf{Summary: }Click 2 sibling final outcomes.} \\
\cmidrule{2-8}
& \multirow{9}{0.5\textwidth}{\textbf{Description: }
\begin{enumerate}
    \item Click on a node satisfying all of the following conditions: \begin{enumerate}[label=-]
        \item it is an unobserved leaf
    \end{enumerate} 
    \item Click on the nodes satisfying all of the following conditions: \begin{enumerate}[label=-]
        \item they are unobserved leaves
    \end{enumerate}
    Click in this way as long as:
    \begin{enumerate}[label=-]
        \item the previously observed node was their sibling.
    \end{enumerate}
    Repeat this step as long as possible.
\end{enumerate}
\textbf{LTL formula: }\texttt{among(is\_leaf and not(is\_observed)) AND NEXT among(is\_leaf and not(is\_observed)) and is\_previous\_observed\_sibling UNTIL IT STOPS APPLYING}}
& 6.1\% & 0.54 & 0.54 & 0.37 & 6 & N\\
& & & & & & & \\
& & & & & & & \\
& & & & & & & \\
& & & & & & & \\
& & & & & & & \\
& & & & & & & \\
& & & & & & & \\
& & & & & & & \\
& & & & & & & \\
& & & & & & & \\
& & & & & & & \\
& & & & & & & \\
& & & & & & & \\
& & & & & & & \\
& & & & & & & \\
\cmidrule{1-8}

\multirow{8}*{21} & \multicolumn{6}{c}{\textbf{Summary: }Random planning.} \\
\cmidrule{2-8}
& \textbf{Description: }
\begin{enumerate}
    \item Terminate or click on some random nodes and then terminate. Repeat this step as long as possible.
\end{enumerate} 
\textbf{LTL formula: }\texttt{True UNTIL IT STOPS APPLYING}
& 4.1\% & 0.57 & 1.0 & 1.0 & 20 & Y\\
\cmidrule{1-8}

\pagebreak

\multirow{38}*{12} & \multicolumn{6}{c}{\textbf{Summary: }Depth first search on two random branches.} \\
\cmidrule{2-8}
& \multirow{9}{0.55\textwidth}{\textbf{Description: }
\begin{enumerate}
    \item Click on a node satisfying all of the following conditions: \begin{enumerate}[label=-]
        \item it is an unobserved node
        \item it has a parent with the lowest value considering the parents of other unobserved nodes.
    \end{enumerate}
    \item Click on the nodes satisfying all of the following conditions: \begin{enumerate}[label=-]
        \item they are unobserved non-roots
        \item they have parents with the highest values considering the parents of other unobserved non-roots.
    \end{enumerate}
    Repeat this step as long as possible.
    \item Click on a node satisfying all of the following conditions: \begin{enumerate}[label=-]
        \item it is an unobserved node
        \item it has a parent with the lowest value considering the parents of other unobserved nodes.
    \end{enumerate}
    \item Click on the nodes satisfying all of the following conditions: \begin{enumerate}[label=-]
        \item they are unobserved non-roots
        \item they have parents with the highest values considering the parents of other unobserved non-roots.
    \end{enumerate}
    Repeat this step as long as possible.
\end{enumerate}
\textbf{LTL formula: }\texttt{among(not(has\_child\_lowest\_level\_value) and not(is\_observed) : has\_parent\_[lowest|highest]\_value) AND NEXT among(not(is\_observed) and not(is\_root) : has\_parent\_highest\_value) UNTIL IT STOPS APPLYING AND NEXT mong(not(has\_child\_lowest\_level\_value) and not(is\_observed) : has\_parent\_[lowest|highest]\_value) AND NEXT among(not(is\_observed) and not(is\_root) : has\_parent\_highest\_value) UNTIL IT STOPS APPLYING}} 
& 3\% & 0.7 & 0.66 & 0.54 & 2 & N\\
& & & & & & & \\
& & & & & & & \\
& & & & & & & \\
& & & & & & & \\
& & & & & & & \\
& & & & & & & \\
& & & & & & & \\
& & & & & & & \\
& & & & & & & \\
& & & & & & & \\
& & & & & & & \\
& & & & & & & \\
& & & & & & & \\
& & & & & & & \\
& & & & & & & \\
& & & & & & & \\
& & & & & & & \\
& & & & & & & \\
& & & & & & & \\
& & & & & & & \\
& & & & & & & \\
& & & & & & & \\
& & & & & & & \\
& & & & & & & \\
& & & & & & & \\
& & & & & & & \\
& & & & & & & \\
& & & & & & & \\
& & & & & & & \\
& & & & & & & \\
& & & & & & & \\
& & & & & & & \\
& & & & & & & \\
& & & & & & & \\
& & & & & & & \\
& & & & & & & \\
& & & & & & & \\
& & & & & & & \\
\cmidrule{1-8}

\pagebreak

\multirow{38}*{17} & \multicolumn{6}{c}{\textbf{Summary: }Depth First Search until a positive value.} \\
\cmidrule{2-8}
& \multirow{9}{0.5\textwidth}{\textbf{Description: }
\begin{enumerate}
    \item Click on a node satisfying all of the following conditions:
    \begin{enumerate}[label=-]
        \item it is an unobserved node
        \item it has a parent with the lowest value considering the parents of other unobserved nodes.
    \end{enumerate}
    \item Click on the nodes satisfying all of the following conditions:
    \begin{enumerate}[label=-]
        \item they are unobserved non-roots
        \item they have parents with the highest values considering the parents of other unobserved non-roots.
    \end{enumerate}
    Repeat this step as long as possible
    \item Click on a node satisfying all of the following conditions:
    \begin{enumerate}[label=-]
        \item it is an unobserved node
        \item it has a parent with the lowest value considering the parents of other unobserved nodes.
    \end{enumerate}
    \item Click on the nodes satisfying all of the following conditions:
    \begin{enumerate}[label=-]
        \item they are unobserved non-roots
        \item they have parents with the highest values considering the parents of other unobserved non-roots.
    \end{enumerate}
    Repeat this step until all the roots are observed or the previously observed node uncovers a positive value
    \item GOTO step 3.
\end{enumerate}
\textbf{LTL formula: }\texttt{among(not(is\_observed) : has\_parent\_lowest\_value) AND NEXT among(not(is\_root) and not(is\_observed) : has\_parent\_highest\_value) UNTIL IT STOPS APPLYING AND NEXT among(not(is\_observed) : has\_parent\_lowest\_value) AND NEXT among(not(is\_root) and not(is\_observed) : has\_parent\_highest\_value) UNTIL (are\_roots\_observed or is\_previous\_observed\_positive)}}
& 2.8\% & 0.85 & 0.68 & 0.47 & 1 & N\\
& & & & & & & \\
& & & & & & & \\
& & & & & & & \\
& & & & & & & \\
& & & & & & & \\
& & & & & & & \\
& & & & & & & \\
& & & & & & & \\
& & & & & & & \\
& & & & & & & \\
& & & & & & & \\
& & & & & & & \\
& & & & & & & \\
& & & & & & & \\
& & & & & & & \\
& & & & & & & \\
& & & & & & & \\
& & & & & & & \\
& & & & & & & \\
& & & & & & & \\
& & & & & & & \\
& & & & & & & \\
& & & & & & & \\
& & & & & & & \\
& & & & & & & \\
& & & & & & & \\
& & & & & & & \\
& & & & & & & \\
& & & & & & & \\
& & & & & & & \\
& & & & & & & \\
& & & & & & & \\
& & & & & & & \\
& & & & & & & \\
& & & & & & & \\
& & & & & & & \\
& & & & & & & \\
& & & & & & & \\
\cmidrule{1-8}

\pagebreak

\multirow{38}*{18} & \multicolumn{6}{c}{\textbf{Summary: }Click sibling final outcomes and a middle node.} \\
\cmidrule{2-8}
& \multirow{9}{0.5\textwidth}{\textbf{Description: }
\begin{enumerate}
    \item Click on random nodes. Do not click on the nodes satisfying either of the following conditions:
    \begin{enumerate}[label=-]
        \item they are non-leaves that have children with the non-highest value on their level.
    \end{enumerate}
    Repeat this step until a node with a positive value is observed or the previously observed node uncovers a -48.
    \item Click on random nodes. Click in this way as long as:
    \begin{enumerate}[label=-]
        \item the previously observed node was their sibling.
    \end{enumerate}
    Do not click on the nodes satisfying either of the following conditions: 
    \begin{enumerate}[label=-]
        \item they are non-leaves that have children with the non-highest value on their level.
    \end{enumerate}
    Repeat this step until a node with a positive value is observed or the termination rewrd is -30.
    \item Click on a node satisfying all of the following conditions: 
    \begin{enumerate}[label=-]
        \item it is a non-leaf that has a child with the non-highest value on its level.
    \end{enumerate}
\end{enumerate}
\textbf{LTL formula: }\texttt{not(among(not(has\_child\_highes\_level\_value) and not(is\_leaf))) UNTIL (is\_positive\_observed or is\_previous\_observed\_min) AND NEXT not(among(not(has\_child\_highest\_level\_value) and not(is\_leaf))) and is\_previous\_observed\_sibling UNTIL (is\_positive\_observed or termination\_reward(-30)) AND NEXT among(not(has\_child\_highest\_level\_value) and not(is\_leaf))}}
& 5.5\% & 0.47 & 0.5 & 0.24 & 1 & Y\\
& & & & & & & \\
& & & & & & & \\
& & & & & & & \\
& & & & & & & \\
& & & & & & & \\
& & & & & & & \\
& & & & & & & \\
& & & & & & & \\
& & & & & & & \\
& & & & & & & \\
& & & & & & & \\
& & & & & & & \\
& & & & & & & \\
& & & & & & & \\
& & & & & & & \\
& & & & & & & \\
& & & & & & & \\
& & & & & & & \\
& & & & & & & \\
& & & & & & & \\
& & & & & & & \\
& & & & & & & \\
& & & & & & & \\
& & & & & & & \\
& & & & & & & \\
& & & & & & & \\
& & & & & & & \\
& & & & & & & \\
& & & & & & & \\
& & & & & & & \\
& & & & & & & \\
& & & & & & & \\
& & & & & & & \\
& & & & & & & \\
& & & & & & & \\
& & & & & & & \\
\cmidrule{1-8}

\pagebreak

\multirow{38}*{19} & \multicolumn{6}{c}{\textbf{Summary: }Depth First Search until a positive valued path.} \\
\cmidrule{2-8}
& \multirow{9}{0.5\textwidth}{\textbf{Description: }
\begin{enumerate}
    \item Click on the nodes satisfying all of the following conditions:
    \begin{enumerate}[label=-]
        \item they are unobserved nodes that lead to leaves whose value is different from -48
        \item they are located on the highest level considering theunobserved nodes that lead to leaves whose value is different from -48.
    \end{enumerate}
    Repeat this step until a node with a positive value is observed or the previously observed node uncovers a -48.
    \item Click on the nodes satisfying all of the following conditions:
    \begin{enumerate}[label=-]
        \item it is an unobserved node that leads to a leaf whose value is different from -48
        \item it is located on the highest level considering the unobserved nodes that lead to leaves whose value is different from -48.
        \item it is the previously observed node was its sibling.
        \end{enumerate}
    \item GOTO step 1 unless all the leaves are observed or all the roots are observed.
\end{enumerate}
\textbf{LTL formula: }\texttt{among(not(has\_leaf\_lowest\_level\_value) and not(is\_observed) : has\_largest\_depth) UNTIL (is\_positive\_observed or is\_previous\_observed\_min) AND NEXT among(not(has\_leaf\_lowest\_level\_value) and not(is\_observed) : has\_largest\_depth) and is\_previous\_observed\_sibling 
\\
LOOP FROM among(not(has\_leaf\_lowest\_level\_value) and not(is\_observed) : has\_largest\_depth) UNLESS (are\_leaves\_observed or are\_roots\_observed)}}
& 3.9\% & 0.82 & 0.52 & 0.41 & 1 & N\\
& & & & & & & \\
& & & & & & & \\
& & & & & & & \\
& & & & & & & \\
& & & & & & & \\
& & & & & & & \\
& & & & & & & \\
& & & & & & & \\
& & & & & & & \\
& & & & & & & \\
& & & & & & & \\
& & & & & & & \\
& & & & & & & \\
& & & & & & & \\
& & & & & & & \\
& & & & & & & \\
& & & & & & & \\
& & & & & & & \\
& & & & & & & \\
& & & & & & & \\
& & & & & & & \\
& & & & & & & \\
& & & & & & & \\
& & & & & & & \\
& & & & & & & \\
& & & & & & & \\
& & & & & & & \\
& & & & & & & \\
& & & & & & & \\
& & & & & & & \\
& & & & & & & \\
& & & & & & & \\
& & & & & & & \\
& & & & & & & \\
& & & & & & & \\
& & & & & & & \\
\cmidrule{1-8}

\bottomrule
\caption{Strategies found in the median combined 10-run of Human-Interpret listed with their ID from the main text. For each strategy, we provide automatically generated descriptions that represent that strategy, and a summary of that strategy that we created by hand. FR denotes the frequency of the strategy; FCF (fit cluster-formula) averages two proportions: formula demonstrations agreeing with the softmax clusters and vice-versa measured using 100000 demonstrations; FON (fit optimal-non-optimal) quantifies how often people's planning operations in the cluster agreed with the description; FPO (fit per operation) is the ratio between the average likelihood per planning operation belonging to the cluster and the average likelihood per planning operation for the (policy induced by the) cluster's description in general; N is the total number of clusters encoding a strategy.}
\label{tab:fullstats}
\end{longtable}

\end{document}